\documentclass{article}
\usepackage{ijcai26}

\usepackage{times}
\usepackage{soul}
\usepackage{url}
\usepackage[hidelinks]{hyperref}
\usepackage[utf8]{inputenc}
\usepackage[small]{caption}
\usepackage{graphicx}
\usepackage{amsmath}
\usepackage{amsthm}
\usepackage{booktabs}
\usepackage{algorithm}
\usepackage{algorithmic}
\usepackage[switch]{lineno}

\usepackage{subcaption}
\usepackage{xspace}
\usepackage{cleveref}
\usepackage{subcaption}

\usepackage{helvet}
\usepackage{courier}
\usepackage{xcolor}

\definecolor{pinegreen}{rgb}{0.0, 0.47, 0.44}
\definecolor{cornellred}{rgb}{0.7, 0.11, 0.11}
\definecolor{cadmiumgreen}{rgb}{0.0, 0.42, 0.24}
\definecolor{royalblue}{rgb}{0.0, 0.14, 0.4}
\definecolor{spirodiscoball}{rgb}{0.06, 0.75, 0.99}
\definecolor{mylightblue}{rgb}{0.85, 0.90, 0.94}
\definecolor{kaistblue}{RGB}{20,135,200}
\definecolor{auburn}{RGB}{166,38,57}

\newcommand{\colorup}[1]{~{\color{cadmiumgreen}(+#1)}}

\newcommand{\down}[1]{~{\color{gray! 30}(\,-#1)}}

\newcommand{\stdv}[1]{{\tiny$\pm$#1}}

\usepackage{amssymb}  
\usepackage{natbib}
\usepackage{adjustbox}

\def\algo{{\textsc{SHED}}\xspace}

\title{\algo: Style-Homogenized Embedding Alignment for Domain Generalization}

\author{
Kai~Gan
\quad
Tong~Wei\thanks{Corresponding author}\\
\affiliations
$^1$School of Computer Science and Engineering, Southeast University, Nanjing 210096, China\\
$^2$Key Laboratory of Computer Network and Information Integration (Southeast University),\\ Ministry of Education, China\\
\emails
\texttt{\{gank, weit\}@seu.edu.cn}
}

\begin{document}

\maketitle

\begin{abstract}
    Domain generalization aims to enhance model robustness against unseen domains with embedding distribution shifts. While large-scale vision-language models like CLIP exhibit strong generalization, their direct image-text embedding alignment suffers from inherent information asymmetry: images encode both class semantics and domain-specific styles, whereas text prompts primarily convey basic class cues. This asymmetry hinders generalization to novel domains in realistic scenarios. To address this, we propose Style-Homogenized Embedding alignment for Domain-generalization (\algo), a novel CLIP-based method that aligns style-homogenized embeddings instead of raw representations from encoders in CLIP. During training, \algo\ removes domain-specific style centroids from both image embeddings computed per source domains and text embeddings which are averaged across diverse prompt templates and stripped of a global centroid. For inference, considering the lack of target domain information, \algo\ projects diverse textual domain centroids into the visual space and aggregates predictions via membership weighting. Extensive experiments on five benchmarks show \algo\ achieves state-of-the-art performance, outperforming prior methods significantly (e.g., +4.0\% on DomainNet vs. standard fine-tuning). 
\end{abstract}

\section{Introduction}

When machine learning models are deployed in real-world open environments, they inevitably encounter samples from unseen domains with embedding distribution shifts. To improve generalization to unseen domains, domain generalization methods \cite{zhou2022domain,li2018domain,li2018deep} aim to learn domain-invariant embeddings by leveraging knowledge from source domains and maintaining robustness across diverse unknown domains. However, models trained from scratch on small-scale datasets often struggle to exhibit strong generalization capabilities across various domains. 

\begin{figure*}[t]
\centering
\begin{subfigure}[b]{0.32\textwidth}
    \includegraphics[width=\textwidth]{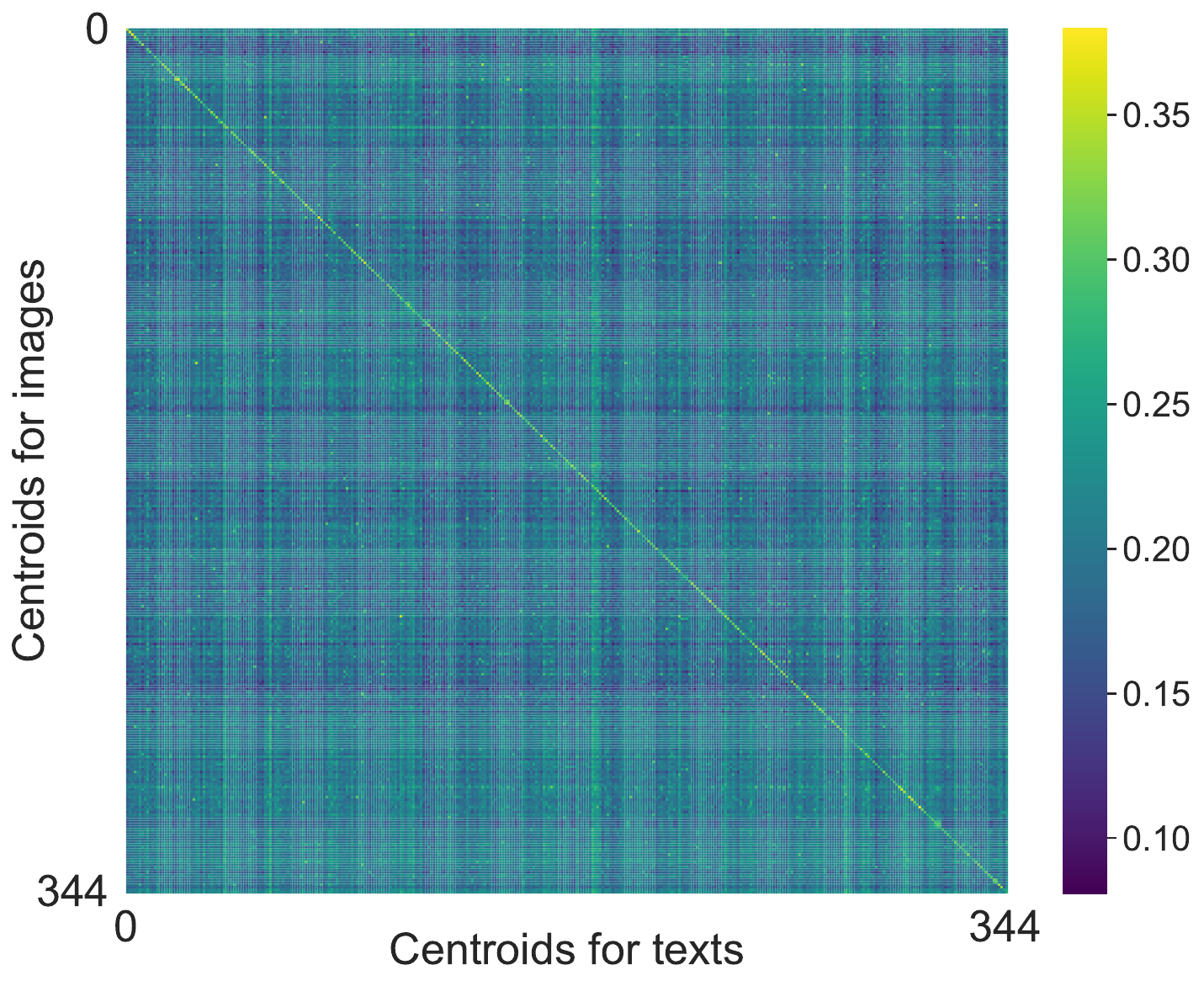}
    \caption{Original CLIP embeddings}
    \label{fig:clip_nodelta}
\end{subfigure}
\hfill
\begin{subfigure}[b]{0.32\textwidth}
    \includegraphics[width=\textwidth]{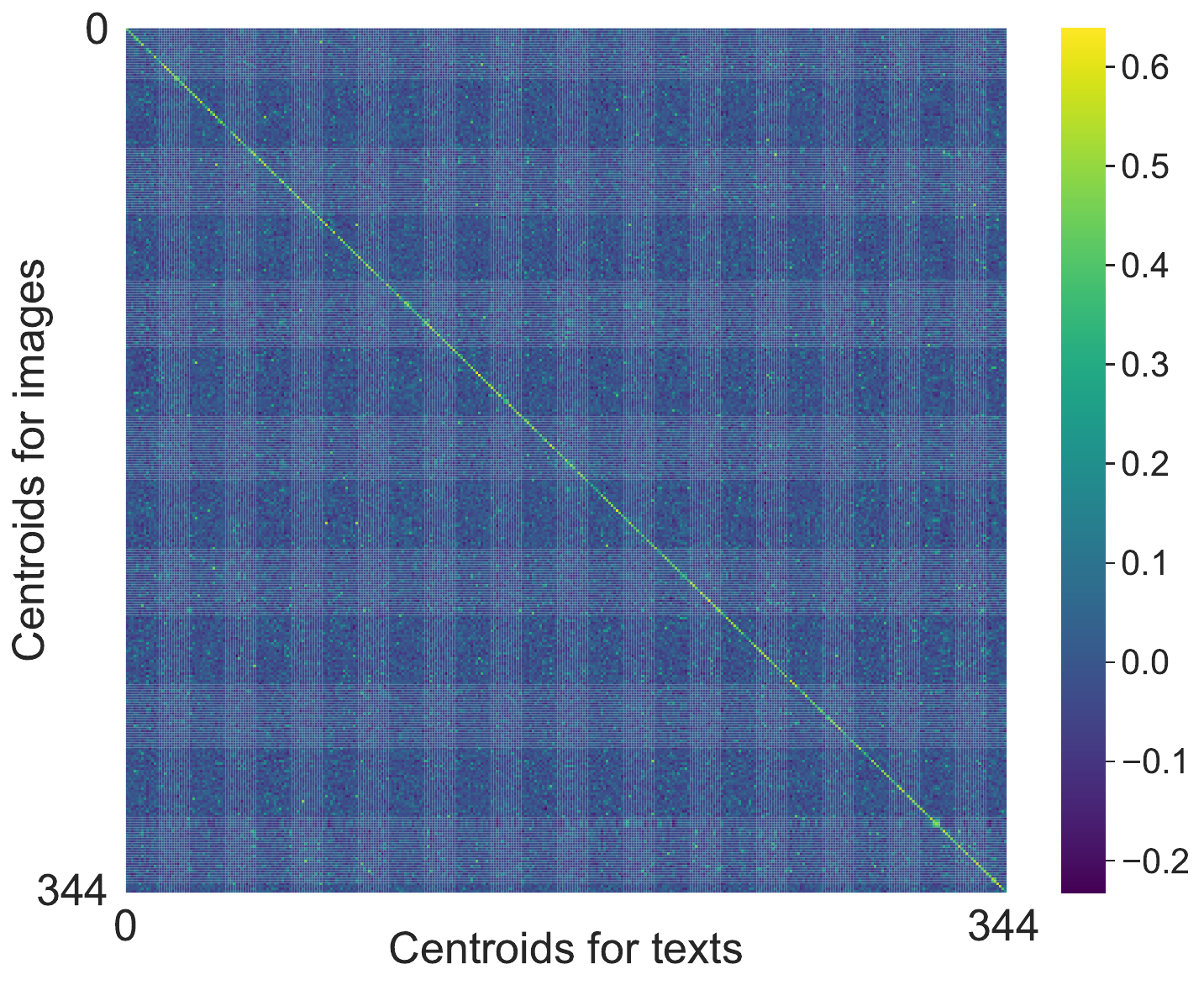}
    \caption{Style-homogenized embeddings}
    \label{fig:clip_delta}
\end{subfigure}
\hfill
\begin{subfigure}[b]{0.32\textwidth}
    \includegraphics[width=\textwidth]{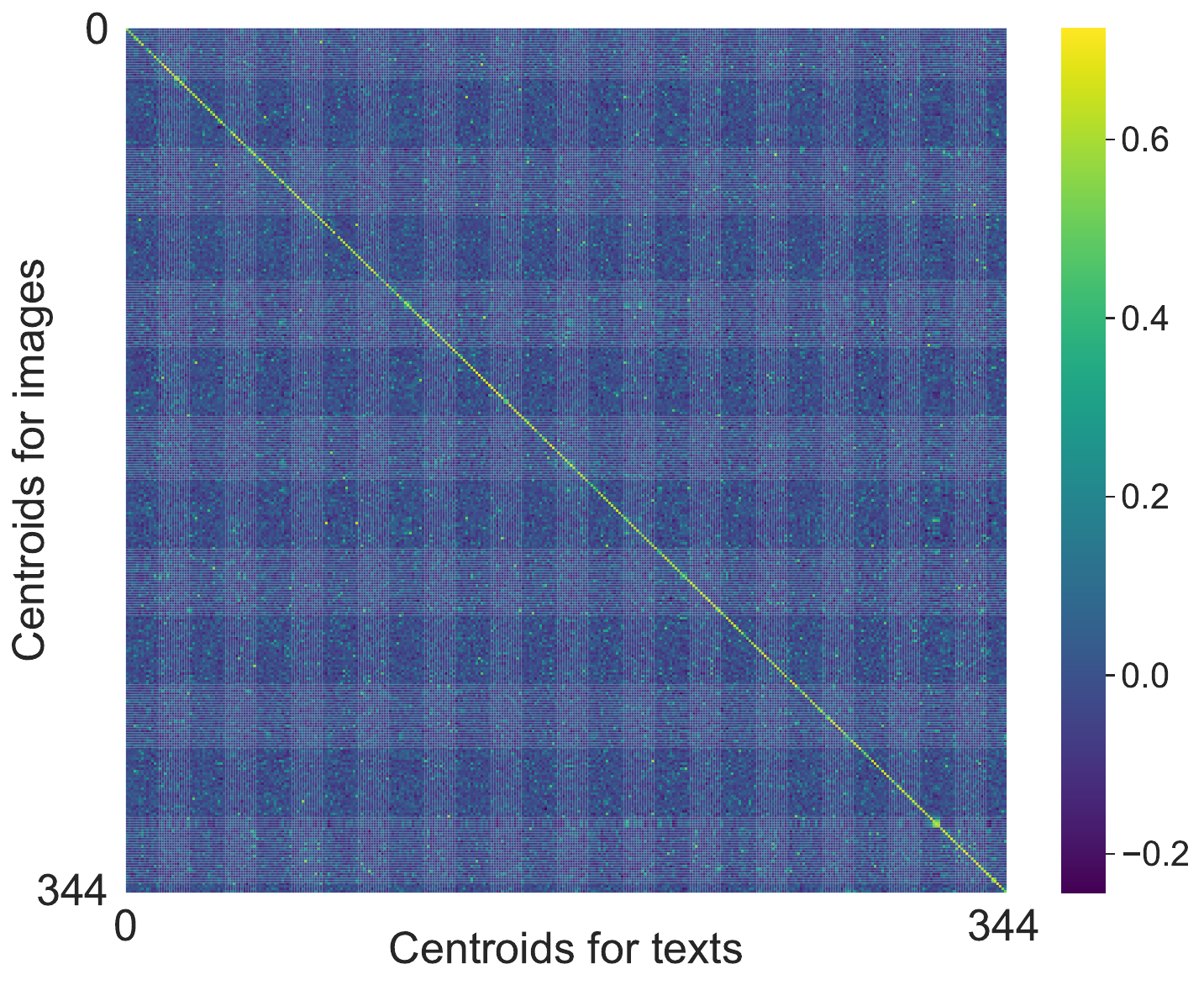}
    \caption{Embeddings after \algo\ training}
    \label{fig:ours_delta}
\end{subfigure}
\caption{The heatmaps of cross-modal centroids similarities for ``Clipart'' domain in DomainNet~\cite{peng2019moment}. For clarity, the ``image centroid'' for each class is defined as the mean embedding of all images within that class, and the ``text centroid'' for each class is defined as the mean embedding derived from multiple domain-styled prompt templates for that class.}

\label{fig:intro}
\end{figure*}

Fortunately, recent advances such as CLIP \cite{radford2021learning}, which is trained on large-scale image-text pairs, have achieved notable success in numerous downstream tasks \cite{li2021align,zhou2022learning,zhou2022conditional,wu2023clipself,gan2024erasing} due to its rich semantic representation and robust zero-shot learning ability. Nevertheless, current mainstream methods predominantly emphasize adaptation to specific domains \cite{lai2023padclip,feng2024rethinking}, which can significantly impair their generalization capability to unseen style domains. To address this challenge, CLIPood \cite{shu2023clipood} introduced margin metric softmax and beta moving average to preserve the original semantic associations of pre-trained CLIP. CLIPCEIL \cite{yu2024clipceil} refined the embedding channels in the visual domain to ensure they contain domain-invariant and class-relevant embeddings by using a lightweight adapter. However, a key limitation of existing methods is their reliance on direct alignment between images embeddings and manually designed class template texts embeddings. Since images encode both class and domain style information while the texts contain only basic class cues, this asymmetric alignment impairs the model’s generalization ability to unseen domains. Domain-specific attributes (e.g., artistic style in paintings) may dominate image representations, causing spurious correlations with text centroids and misleading predictions. Specifically, as illustrated in \Cref{fig:clip_nodelta}, we observe that in the original CLIP embedding space, each image centroid exhibits high similarity with multiple class text centroids, which makes it difficult to infer the true class during inference accurately. This naturally raises the question: \textit{Can we achieve symmetric alignment training between images and text embeddings?}

\begin{figure}[t]
\centering
\includegraphics[width=\columnwidth]{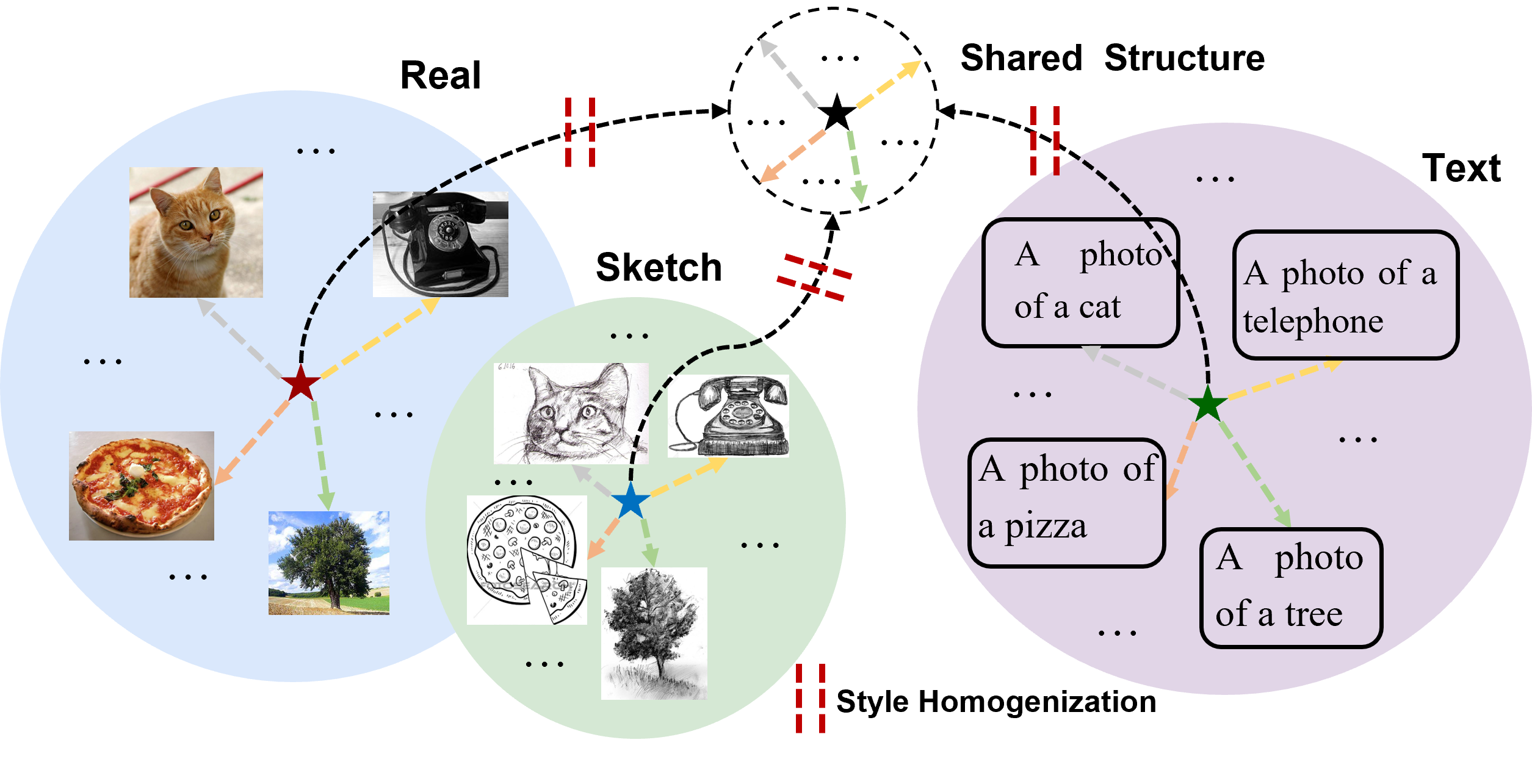}
\caption{An illustration of the embedding space distribution for image samples from different domains and class texts constructed using templates. The red dashed line indicates the gap between the image and text modalities.}
\label{fig:delta_structure}
\end{figure}

In this paper, we propose a novel CLIP-based domain generalization method \algo\ to answer this question. Instead of aligning the raw image-text embeddings directly, \algo\ aligns the embeddings after removing domain-specific style information. As visualized in \Cref{fig:delta_structure}, \algo\ exploits a key geometric insight: embeddings across various domains and template-based texts share a latent structural consistency where samples from the same class consistently point in similar directions relative to their respective domain or text centroids. This structure consistency motivates us to explicitly performing style-homogenized embedding alignment within this shared geometric structure. Concretely, for images, we derive style-homogenized image embeddings by first computing the centroid of each domain, and then subtracting it from the corresponding sample embeddings. For texts, we generate text embeddings using various templates and average them to derive a representative text embedding for each class. We then calculate a global text centroid across all classes and subtract it from each class embedding to obtain style-homogenized text embeddings. As shown in \Cref{fig:clip_nodelta,fig:clip_delta}, the style-homogenized image and text embeddings exhibit more discriminative inter-class separability compared to the original CLIP embeddings, which facilitates more accurate and robust classification \cite{shi2024long} across different domains. Unsurprisingly, when using style-homogenized image and text embeddings to calculate zero-shot style-homogenized probability, we find it consistently achieve superior zero-shot classification performance across various domains compared to using original CLIP embeddings in \Cref{fig:delta_acc}.

During inference, it is challenging to obtain style-homogenized image embeddings due to the domain identity of images are unavailable. To tackle this issue, we manually construct multiple textual domain centroids and project them into the visual space to approximate potential image domain centroids. Accordingly, we compute the membership degree of the test samples to each visual centroid, which serves as weights for aggregating the style-homogenized probabilities from these centroids. As a result, \algo\ eliminates dependence on ground-truth domain centroids and can predict style-homogenized probabilities for arbitrary samples.

Our main contributions can be summarized as follows:
\begin{itemize}
\item We propose \algo, a novel domain generalization framework that enhances prediction on unseen domains. It uniquely addresses the inherent information asymmetry in CLIP by aligning style-homogenized embeddings, rather than performing direct alignment on raw visual and textual representations.
\item We propose the aggregation of multiple centroids predictions to equip the model with the capability to handle samples from any target domains during inference.
\item Extensive experimental analyses show that \algo achieves state-of-the-art performance on multiple benchmarks across various domains, e.g., \algo improves ERM by over 4.0\% on DomainNet.
\end{itemize}

\section{Related Work}

\paragraph{Vision-Language Models} Recently, vision-language pre-training \cite{radford2021learning,jia2021scaling,li2022blip,li2023blip,pham2023combined,luo2025lada,gan2025semi,wei2026x,wei2024vision} have demonstrated remarkable success in various downstream tasks through fine-tuning. Specifically, CLIP \cite{radford2021learning} leverages contrastive learning on a large collection of image-text pairs to learn semantic associations between visual and textual modalities, enabling cross-modal retrieval and classification. BLIP \cite{li2022blip} integrates vision-language matching and generation tasks with guided objectives to enhance image-text alignment, enabling strong performance on both vision-language understanding and generation tasks. BLIP-2 \cite{li2023blip} introduces a lightweight query transformer between the vision encoder and the language model to facilitate efficient image-text alignment. Additional, vision-language models models can achieve significant performance improvements on downstream tasks through various fine-tuning strategies \cite{houlsby2019parameter,chen2022adaptformer,jia2022visual}, including long-tailed learning \cite{dong2022lpt,shi2024long}, test-time adaptation \cite{zhou2023ods,osowiechi2024watt}and domain generalization \cite{shu2023clipood,yu2024clipceil}. To address the challenge of domain generalization, we investigate how CLIP's robust vision-language pretraining can be adapted to enhance model robustness across unseen domains.

\paragraph{Domain Generalization} The goal of domain generalization is to train models on data from multiple known domains, while evaluating them on data from entirely unseen domains. Specifically, SagNets \cite{nam2021reducing} improve generalization and adaptability in cross-domain tasks by disentangling style and category representations, thereby reducing the CNN's reliance on style-biased embeddings. BNE \cite{segu2023batch} employs domain-specific batch normalization layers to extract domain-dependent embeddings and models each domain in a shared latent space based on distance. mDSDI \cite{bui2021exploiting} operates by decomposing latent embeddings into domain-invariant and domain-specific components. It jointly learns representations and employs meta-learning to optimize the domain-specific part. However, existing methods primarily rely on direct alignment of image and text embeddings, neglecting the inherent information asymmetry between the two modalities. In contrast, our work is the first to systematically address this asymmetry by proposing a symmetric alignment of style-homogenized embeddings.


\paragraph{Domain Generalization via CLIP}

Considering the strong semantic generalization capabilities of vision-language models such as CLIP, recent studies have begun to explore how to leverage CLIP to improve performance on domain generalization tasks. MIRO \cite{cha2022domain} introduces a pretrained model as an approximation to the oracle model and maximizes the mutual information with it to encourage the learning of more generalizable representations. VL2V-ADiP \cite{addepalli2024leveraging} enhances the generalization ability of the student model by aligning the visual and language modalities of the teacher VLM with the visual modality of the student, followed by distillation of the aligned representations, while preserving the student's pretrained characteristics. CLIPood \cite{shu2023clipood} introduces a Margin Metric Softmax loss with class-adaptive boundaries to fine-tune the model by leveraging semantic relationships in the text modality, while employing a Beta distribution-based moving average to fuse the pretrained zero-shot model and the fine-tuned model. However, existing methods primarily rely on direct alignment of image and text embeddings, neglecting the inherent information asymmetry between the two modalities.

\section{The Proposed Method}

In this section,  we provide a detailed description of \algo\, which includes style-homogenized alignment during model training and domain-agnostic centroid aggregation for unknown domain sample inference.

\subsection{Preliminaries}

In this paper, we investigate the problem of domain generalization. Formally, our objective is to train the model on the source training set $\{\mathcal{D}^s = \{(x_i^{s}, y_i^{s})\}^{n_s}_{i=1}\}^{S}_{s=1}$, where $S$ represents the number of domains and $n_s$ denotes the number of samples in each source domain. $y_i^{s} \in \{0, 1\}^{C}$ is the ground truth class label of $C$-class classification task for image $x_i^{s}$. During the testing phase, we evaluate the model on a novel domain dataset $\mathcal{D}^{\mathrm{test}} = \{x_j\}^{n_{test}}_{j=1}$ that was not observed during training.

We focus on generalizing the vision-language pre-trained model CLIP \cite{radford2021learning} to unseen domains. During training, we fine-tune only the visual encoder $f_{I}$, while the CLIP text encoder remains frozen with class name text embeddings $\{T_c\}^{C}_{c=1}$ pre-encoded using prompt templates (e.g., ``\texttt{a photo of a \{class name\}}''). This allows us to predict the class probabilities of the image $x$:
\begin{equation}
    \mathcal{P}(y|x) = \frac{\exp(\langle f_I(x), T_y\rangle / \tau)}{\sum^{C}_{c=1} \exp(\langle f_I(x), T_c\rangle / \tau))}
    \label{eq:clip_prob}
\end{equation}
where $\langle \cdot, \cdot \rangle$ represents the cosine similarity for normalized embeddings and $\tau$ is a temperature scaler. $f_I(x) \in \mathbb{R}^d$ is the image embedding for sample $x$.

\subsection{Style-homogenized Alignment}

Existing fine-tuning strategies for CLIP typically rely on direct alignment between image and text embeddings. However, such alignment may be mismatched due to extra domain-specific information encoded in the visual embeddings but not in the text embeddings, which can degrade the model’s ability to generalize to unseen domains. Accordingly, \algo\ focuses on aligning style-homogenized embeddings between image and text during fine-tuning:
\begin{equation}
    \mathcal{L}_c = - \log \frac{\exp(\langle \hat{f}_I(x) \cdot \hat{T}_y\rangle / \tau)}{\sum^{C}_{c=1} \exp(\langle \hat{f}_I(x) \cdot \hat{T}_c\rangle / \tau))}
    \label{eq:delta_align}
\end{equation}
where $\hat{f}_I(x) \in \mathbb{R}^d$ and $\hat{T} \in \mathbb{R}^d$ represent the style-homogenized embeddings for images and texts. 

Specifically, $\hat{f}_I(x)$ is computed by removing the domain centroid from the original image embedding:
\begin{equation}
\begin{gathered}
    \hat{f}_I(x) = \frac{f_I(x) - \mu_{s}}{\| f_I(x) - \mu_{s} \|}, \;\;
    \mu_s = \frac{1}{n_s} \sum_{x \in \mathcal{D}^s} f_I(x)
    \label{eq:delta_image}
\end{gathered}
\end{equation}
where $\mu_s$ represents the centroid for domain $s$. It is obvious that the domain centroid captures the most concentrated or representative information of the domain \cite{chen2023center,liangumfc}. When the domain centroid is removed from sample embeddings, the remaining style-homogenized embeddings are expected to preserve mainly class-specific information. It is worth noting that $\mu_s$ is computed prior to training via some passes over the training dataset, and it remains unchanged throughout the subsequent training. We argue that, since the initial model embeddings retain the semantic information of the original CLIP, fixing $\mu_s$ helps preserve CLIP’s strong generalization capability during later training.

Similarly, we apply templates $\{T_s\}^{S}_{s=1}$ (e.g., ``\texttt{a \{domain name\} photo of \{class name\}}'', where domain styles are known during training such as ``drawing'' and ``cartoon'') to each class and perform aggregation:
%
\begin{equation}
\begin{gathered}
    T_c = \frac{1}{S} \sum^{S}_{s=1} f_T(T_s(c)),
    \label{eq:T_c}
\end{gathered}
\end{equation}
where $f_T$ is the frozen text encoder from CLIP model. Then we can obtain the domain-invariant text embeddings:
\begin{equation}
\begin{gathered}
    \hat{T}_c = \frac{T_c - \mu^{\mathrm{text}}}{\| T_c - \mu^{\mathrm{text}} \|},\;\;
    \mu^{\mathrm{text}} = \frac{1}{C} \sum^{C}_{c=1} T_c.
    \label{eq:delta_text}
\end{gathered}
\end{equation}

As shown in \Cref{fig:clip_nodelta}, $\hat{f}_I(x) \in \mathbb{R}^d$ and $\hat{T}_c$ exhibit high similarity for their corresponding classes, indicating that style-homogenized embeddings are highly discriminative. Intuitively, subtracting centroids facilitates every domain to share a common structure, making cross-modal directions comparable. By aligning the model via \Cref{eq:delta_align} during training, the discriminative power of style-homogenized embeddings can be further improved as illustrated by subtle deeper color in \Cref{fig:ours_delta} compared to \Cref{fig:clip_delta}. In our opinion, the minor differences between the two figures indicates that \algo\ largely preserves the powerful semantic representations of the original CLIP model, which facilitates generalization across broader domains.

Additionally, to ensure training stability and further retain the inherent generalization strength of CLIP, \algo\ regularizes the model by encouraging its embeddings to approximate the original CLIP embeddings:
\begin{equation}
\begin{gathered}
    \mathcal{L}_{reg} = \sum^{d}_{i=1} \| f_I(x) - f^{\scriptscriptstyle\mathrm{CLIP}}_I(x) \| + \| \hat{f}_I(x) - \hat{f}^{\scriptscriptstyle\mathrm{CLIP}}_I(x) \| ,
    \label{eq:clip_feats}
\end{gathered}
\end{equation}
where the superscript ``CLIP'' indicates that the embedding is produced by the original CLIP model, and $d$ is explicitly defined as the dimension of the embedding. $\mathcal{L}_{reg}$ employs $L1$ loss to simultaneously constrain both visual embeddings and style-homogenized embeddings. Its effect will be demonstrated in the ablation study.
In summary, the total training objective is $\mathcal{L} = \mathcal{L}_{c} + \mathcal{L}_{reg}$.

\begin{table*}[t]
\centering\small
\vspace{-0.08in}
\begin{adjustbox}{width=\linewidth}
\begin{tabular}{lcccccccc}
\toprule
Method & Venue & PACS & VLCS & OfficeHome & TerraInc & DomainNet & Avg \\
\midrule
CLIP Zero-Shot & ICML'21 & 
96.2 & 81.7 & 82.0 & 33.4 & 57.5 & 70.2 \\
Style-Homogenized Zero-Shot & - &
96.5 & 79.7 & 84.6 & 34.2 & 60.0 & 71.0 \\
ERM & - &
96.1\stdv{0.5} & 83.0\stdv{0.2} & 83.3\stdv{0.3} & 60.9\stdv{0.2} & 59.9\stdv{0.1} & 76.7\stdv{0.2} \\
\midrule
CoOp~\citep{zhou2022learning} & IJCV'22 &
96.0\down{0.1} & 81.1\down{1.9} & 83.5\colorup{0.2} & 47.0\down{13.9} & 59.8\down{0.1} & 73.5\down{3.2} \\
CoCoOp~\citep{zhou2022conditional} & CVPR'22 &
95.7\down{0.4} & 83.1\colorup{0.1} & 84.3\colorup{1.0} & 50.4\down{10.5} & 60.0\colorup{0.1} & 74.7\down{2.0} \\
MIRO~\citep{cha2022domain} & ECCV'22 &
95.6\down{0.5} & 82.2\down{0.8} & 82.5\down{0.8} & 54.3\down{6.6} & 54.0\down{5.9} & 73.7\down{3.0} \\
CLIPood~\citep{shu2023clipood} & ICML'23 &
97.3\stdv{0.1}\colorup{1.2} & 85.0\stdv{0.4}\colorup{2.0} & 87.0\stdv{0.2}\colorup{3.7} & 60.4\stdv{0.7}\down{0.5} & 63.5\stdv{0.1}\colorup{3.6} & 78.6\stdv{0.2}\colorup{1.9} \\
VLV2-SD~\citep{addepalli2024leveraging} & CVPR'24 &
96.7\colorup{0.6} & 83.3\colorup{0.3} & 87.4\colorup{4.1} & 58.5\down{2.4} & 62.8\colorup{2.9} & 77.7\colorup{1.0} \\
CLIPCEIL~\citep{yu2024clipceil} & NeurIPS'24 &
97.2\stdv{0.1}\colorup{1.1} & 85.2\stdv{0.5}\colorup{2.2} & 87.7\stdv{0.3}\colorup{4.4} & 62.0\stdv{0.5}\colorup{1.1} & 63.6\stdv{0.2}\colorup{3.7} & 79.1\stdv{0.2}\colorup{2.4} \\
Diverse Text Prompts~\citep{wen2025domain} & CVPR'25 &
97.0\colorup{0.9} & 84.8\colorup{1.8} & 87.7\colorup{4.4} & 63.3\colorup{2.4} & 63.1\colorup{3.2} & 79.2\colorup{2.5} \\
\algo\ (ours) & - &
\textbf{97.6}\stdv{0.1}\colorup{1.5} & \textbf{85.4}\stdv{0.2}\colorup{2.4} & \,\>\textbf{87.7}\stdv{0.1}\colorup{4.4} & \textbf{62.5}\stdv{0.3}\colorup{1.6} & \textbf{63.9}\stdv{0.1}\colorup{4.0} & \textbf{79.4}\stdv{0.2}\colorup{2.7} \\
\bottomrule
\end{tabular}
\end{adjustbox}
\caption{%
Comparison of classification accuracy on domain generalization benchmarks. Parentheses indicate the performance gap from the standard fine-tuning of CLIP (ERM), where values highlighted in {\color{cadmiumgreen}green} and {\color{gray}gray} indicate performance is better or worse than ERM. The best results are in \textbf{bold}.
}\label{tab:main_res}
\end{table*}

\subsection{Domain-Agnostic Centroid Aggregation}

Although style-homogenized alignment enhances model's ability to distinguish between classes, it is impractical to directly rely on style-homogenized embeddings during inference, as the domain of each test sample cannot be assumed. To solve this challenge, we propose to construct new textual domain centroids of multiple additional domain styles:
\begin{equation}
\begin{gathered}
    \mu_{t} = \frac{1}{C} \sum^{C}_{c=1} f_T(T_t(c)),
    \label{eq:text_centroids}
\end{gathered}
\end{equation}
where $\{T_t\}^{N_T}_{t=1}$ is the templates for additional $N_T$ domain styles and $\mu_{t}$ represents the centroid for template $T_t$. Additional domain styles are provided in the supplementary material. Considering the existing gap \cite{liang2022mind} between visual and textual modalities in CLIP, we adopt two distinct approaches to project the corresponding textual centroids into the visual embedding space.
\begin{itemize}
    \item \textbf{Centroids Projection Method (CPM)}. CPM computes the relative difference between the additional text centroids and the S source domain centroids, and transfers this difference to the visual domain centroids $\mu_s$:
\begin{equation}
\begin{gathered}
    \mu^{\scriptscriptstyle\mathrm{CPM}}_{t} = \frac{1}{S} \sum^{S}_{s=1} \mu_s + (\mu_{t} - \mu^{\mathrm{text}}_s),
    \label{eq:text2image_CPM}
\end{gathered}
\end{equation}
where $\mu^{\scriptscriptstyle\mathrm{text}}_s = \frac{1}{C} \sum^{C}_{c=1} f_T(T_s(c)$ represents the centroids for source domain $s$.
    \item \textbf{Sample Weighting Method (SWM)}. SWM utilizes similarity between new text centroids and visual sample embeddings as weights, and derives visual centroids of additional styles by aggregating weighted visual embeddings:
\begin{equation}
\begin{gathered}
    \mu^{\scriptscriptstyle\mathrm{SWM}}_{t} = \left( \mathrm{softmax}\left(\frac{\mu_t \cdot V^{\scriptscriptstyle\mathrm{CLIP}}}{\tau_{\scriptscriptstyle\mathrm{SWM}}}\right) \right) \cdot V^{\scriptscriptstyle\mathrm{CLIP}},
    \label{eq:text2image_SWM}
\end{gathered}
\end{equation}
where temperature parameter $\tau_{\scriptscriptstyle\mathrm{SWM}}$ is set to a default value of $\frac{1}{100}$. Notably, $V^{\scriptscriptstyle\mathrm{CLIP}}$ is the visual embeddings sampled before training process begins.
\end{itemize}


\begin{table*}[t]
\centering\small
\vspace{-0.08in}
\begin{adjustbox}{width=\linewidth}
\begin{tabular}{@{}lccccccc@{}}
    \toprule
    Ablations &
      Clipart &
      Infograph &
      Painting &
      Quickdraw &
      Real &
      Sketch &
      Avg \\ \midrule
    \algo &
      78.2 &
      54.8 &
      72.9 &
      21.3 &
      85.8 &
      70.4 &
      63.9 \\
    w/o Style-homogenized alignment &
      76.9\down{1.3} &
      53.8\down{1.0} &
      71.4\down{1.5} &
      21.1\down{0.2} &
      85.6\down{0.2} &
      69.0\down{1.4} &
      63.0\down{0.9} \\
    w/o Regularization &
      78.6\colorup{0.4} &
      53.9\down{0.9} &
      72.4\down{0.5} &
      20.3\down{1.0} &
      85.6\down{0.2} &
      69.6\down{0.8} &
      63.4\down{0.5} \\
    w/o Combined predictions &
      78.2\down{0.0} &
      54.5\down{0.3} &
      72.5\down{0.4} &
      21.1\down{0.2} &
      84.6\down{1.2} &
      70.1\down{0.3} &
      63.5\down{0.4} \\
    w/o Additional centroids &
      78.2\down{0.0} &
      54.5\down{0.3} &
      72.5\down{0.4} &
      20.6\down{0.7} &
      85.7\down{0.1} &
      70.1\down{0.3} &
      63.6\down{0.3} \\
    w/o CPM &
      78.3\colorup{0.1} &
      54.6\down{0.2} &
      72.8\down{0.1} &
      20.6\down{0.7} &
      85.7\down{0.1} &
      70.3\down{0.1} &
      63.7\down{0.2} \\
    w/o SWM &
      78.1\down{0.1} &
      54.2\down{0.6} &
      72.7\down{0.2} &
      21.3\down{0.0} &
      85.7\down{0.1} &
      70.3\down{0.1} &
      63.7\down{0.2} \\
    \bottomrule
\end{tabular}
\end{adjustbox}
\caption{Ablation studies. We investigate the impact of the core components of the \algo\ on DomainNet. ``w/o Style-homogenized alignment'' represents that \Cref{eq:delta_align} is replaced by direct alignment of image and text features. The regularization denotes the loss in \Cref{eq:clip_feats}. Parentheses indicate the performance gap from \algo, where values highlighted in {\color{cadmiumgreen}green} and {\color{gray}gray} indicate performance is better or worse than \algo, respectively.}
\label{tab:ablations}
\end{table*}

\begin{figure}[t]
\centering
\includegraphics[width=\columnwidth]{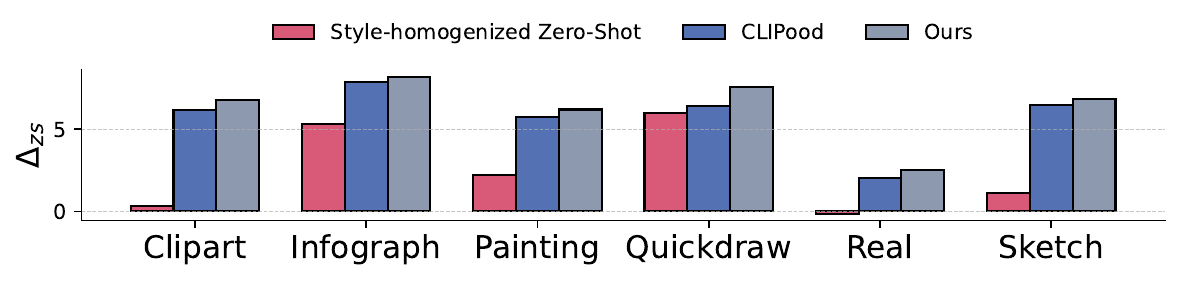}
\caption{Performance change of different methods relative to the zero-shot CLIP  baseline under various domains on DomainNet.}
\label{fig:delta_acc}
\end{figure}

\begin{algorithm}
\caption{\algo\ Training}
\begin{algorithmic}[1]
\REQUIRE Source domains $\mathcal{D}^s$ for $s=1$ to $S$, number of iterations $E$
\REQUIRE Pre-trained CLIP encoders $f_I$, $f_T$

\STATE Compute centroid $\mu_s \leftarrow \text{mean}(f_I^{\text{CLIP}}(x))$ for each domain $s$
\STATE $T_c \gets \frac{1}{S}\sum_{s=1}^{S} f_T(\text{``a [domain] photo of a \{c\}''})$
\STATE Compute global text centroid $\mu^{\text{text}} \leftarrow \text{mean}(T_c)$
\STATE $\hat{T}_c \leftarrow \text{Normalize}(T_c - \mu^{\text{text}})$

\FOR{$e=1$ {\bfseries to} $E$}
    \STATE Sample batch $(x, y, s)$
    \STATE Extract $f_I(x)$
    \STATE $\hat{f}_I(x) \leftarrow \text{normalize}(f_I(x) - \mu_s)$
    \STATE Compute $\mathcal{L}_c$ through \Cref{eq:delta_align}
    \STATE Compute $\mathcal{L}_{reg}$ through \Cref{eq:clip_feats}
    \STATE Update $f_I$ using $\nabla(\mathcal{L}_c + \mathcal{L}_{reg})$
\ENDFOR
\end{algorithmic}
\label{alg:training}
\end{algorithm}

We posit that these two complementary projection methods can bridge the modality gap robustly. CPM operates on a global level, assuming a consistent structural relationship by transferring a relative ``style vector'' from the text space to the visual space. In contrast, SWM takes a more localized, data-driven method, constructing a visual style centroid by attending to source images that are semantically closest to the target style description. By combining both, \algo\ synthesizes a richer and more accurate set of potential visual domain centroids for inference. 

After obtaining a sufficient number of visual domain centroids, the next step is to assign domain centroids for each test sample $x_j \in \mathcal{D}^{\mathrm{test}}$ in inference. \algo\ first combines all visual centroids, i.e., $\mu^{V} = \mathrm{concat}(\{\mu_s\}^{S}_{s=1}, \{\mu^{\scriptscriptstyle\mathrm{CPM}}_{t}\}^{N_T}_{t=1}, \{\mu^{\scriptscriptstyle\mathrm{SWM}}_{t}\}^{N_T}_{t=1})$, and then computes the soft assignment of $f^{\scriptscriptstyle\mathrm{CLIP}}_I(x_j)$ to all centroids:
\begin{equation}
\begin{gathered}
    \pi_j = \mathrm{softmax}\left(\frac{f^{\scriptscriptstyle\mathrm{CLIP}}_I(x_j) \cdot \mu^{V}}{\tau_{\scriptscriptstyle{C}}}\right),
    \label{eq:assign}
\end{gathered}
\end{equation}
where $\tau_{\scriptscriptstyle{C}}$ is the temperature for centroids assignment. It is evident that $\pi_j$ places greater emphasis on centroids more aligned with the style of $x_j$ and can serve as a membership weighting for each domain centroid during inference.

\begin{algorithm}
\caption{\algo\ Inference}
\begin{algorithmic}[1]
\REQUIRE Test sample $x_j$
\REQUIRE style-homogenized text embeddings $\{\hat{T}_c\}$

\STATE \textbf{Construct visual centroids:}
\FOR{each domain style $t$}
    \STATE $\mu_t^{\text{CPM}} \leftarrow \text{project via CPM}$
    \STATE $\mu_t^{\text{SWM}} \leftarrow \text{project via SWM}$
\ENDFOR
\STATE $\mu^V \leftarrow \{\mu_s\} \cup \{\mu_t^{\text{CPM}}\} \cup \{\mu_t^{\text{SWM}}\}$

\STATE \textbf{Compute predictions:}
\STATE $\pi_j \leftarrow \text{softmax}(f_I^{\text{CLIP}}(x_j) \cdot \mu^V)$
\FOR{each centroid $\mu_v \in \mu^V$}
    \STATE $\hat{f}_I \leftarrow \text{normalize}(f_I(x_j) - \mu_v)$
    \STATE $\mathcal{P}_v(y|x_j) \leftarrow \text{softmax}(\langle \hat{f}_I, \hat{T}_y \rangle)$
\ENDFOR
\STATE $\mathcal{P}^{\mu} \leftarrow \sum_v \pi_j(v) \cdot \mathcal{P}_v(y|x_j)$
\STATE \textbf{Final prediction:} $\mathcal{P} \leftarrow \lambda \cdot \mathcal{P}^{\text{CLIP}} + (1-\lambda) \cdot \mathcal{P}^{\mu}$
\RETURN $\arg\max_y \mathcal{P}(y|x_j)$
\end{algorithmic}
\label{alg:inference}
\end{algorithm}

Subsequently, sample $x_j$ utilizes each element in $\mu^{V} = \{\mu^{V}_{v}\}^{2N_T + S}_{v=1}$ as domain centroid to compute the corresponding prediction $\mathcal{P}_v(y|x_j, \mu^{V}_{v})$ following the probability inside the negative logarithm in \Cref{eq:delta_align} and integrate the predictions using soft assignment from $\pi_j$:

\begin{equation}
\begin{gathered}
    \mathcal{P}^{\mu}(y|x_j) = \sum^{2N_T + S}_{v=1} \pi_j(v) \cdot \mathcal{P}_v(y|x_j, \mu^{V}_{v}).
    \label{eq:delta_pred}
\end{gathered}
\end{equation}

Meanwhile, considering the strong zero-shot capability of CLIP, we combine the inference probabilities $\mathcal{P}(y|x_j)$ with its zero-shot prediction $\mathcal{P}^{\scriptscriptstyle\mathrm{CLIP}}(y|x_j)$:
\begin{equation}
\begin{gathered}
    \mathcal{P}(y|x_j) = \mathcal{P}^{\scriptscriptstyle\mathrm{CLIP}}(y|x_j) \cdot \lambda(x_j) + \mathcal{P}^{\mu}(y|x_j) \cdot (1.0 - \lambda(x_j)),
    \label{eq:final_pred}
\end{gathered}
\end{equation}
where $\lambda(x_j) = \max_{1 \leq c \leq C} \mathcal{P}^{\scriptscriptstyle\mathrm{CLIP}}(c \mid x)$ is the confidence of the zero-shot CLIP prediction for $x_j$. When zero-shot CLIP is confident in its prediction, the final decision is primarily guided by CLIP; otherwise, it is dominated by $\mathcal{P}^{\mu}(y|x_j)$. In summary, the overall training and inference procedures of \algo\ are presented in \Cref{alg:training} and \Cref{alg:inference} respectively. 


\section{Experiments}

In this section, we demonstrate that the proposed \algo\ method achieves consistent improvements on domain generalization benchmarks compared to previous methods. We also provide an in-depth analysis of the method and conduct ablation studies to investigate the contribution of each component to overall performance.

\begin{figure}[t]
\centering
\includegraphics[width=\columnwidth]{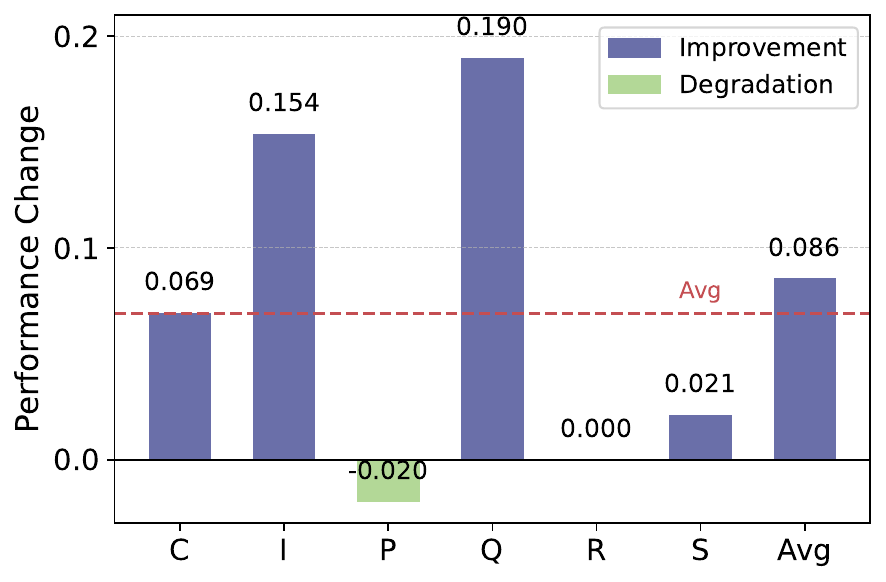}
\caption{Performance change by incorporating additional centroids derived from real target domain style words into \algo. ``C'', ``I'', ``P'', ``Q'', ``R'', and ``S'' represent the first letters of the six target domains in the DomainNet.}
\label{fig:domain_word}
\end{figure}

\begin{figure*}[t]
\centering
\begin{subfigure}[b]{0.245\textwidth}
    \includegraphics[width=\textwidth]{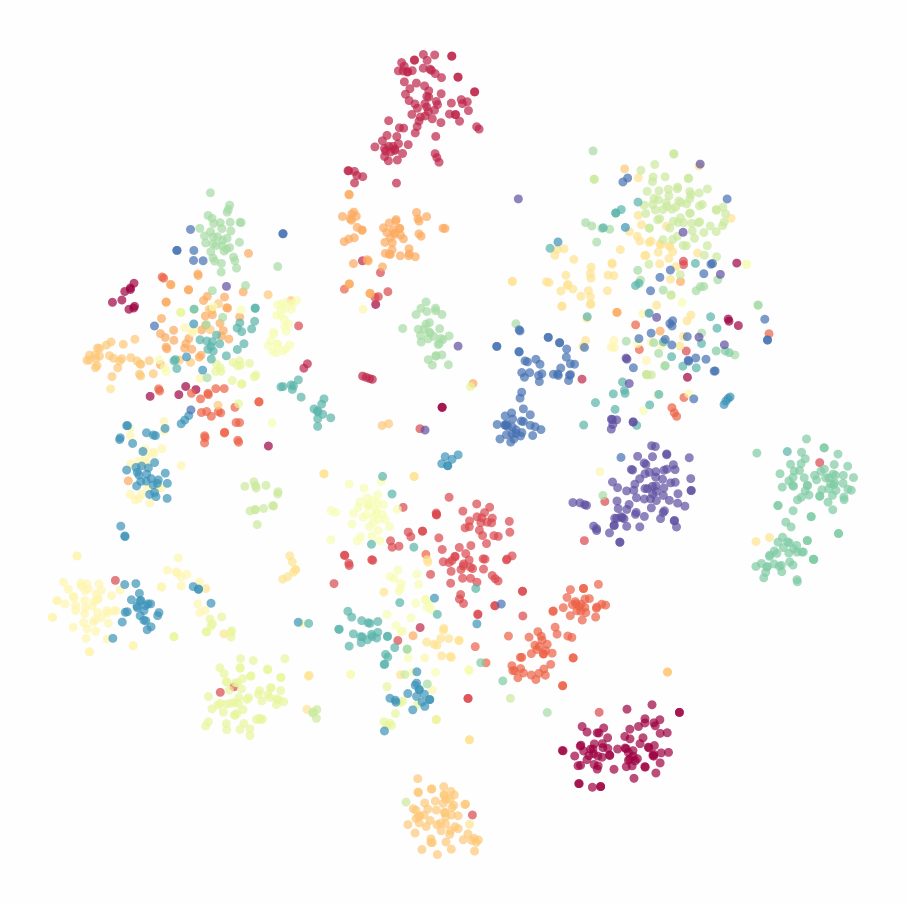}
    \caption{Original CLIP embeddings}
    \label{fig:tsne_clip}
\end{subfigure}
\hfill
\begin{subfigure}[b]{0.245\textwidth}
    \includegraphics[width=\textwidth]{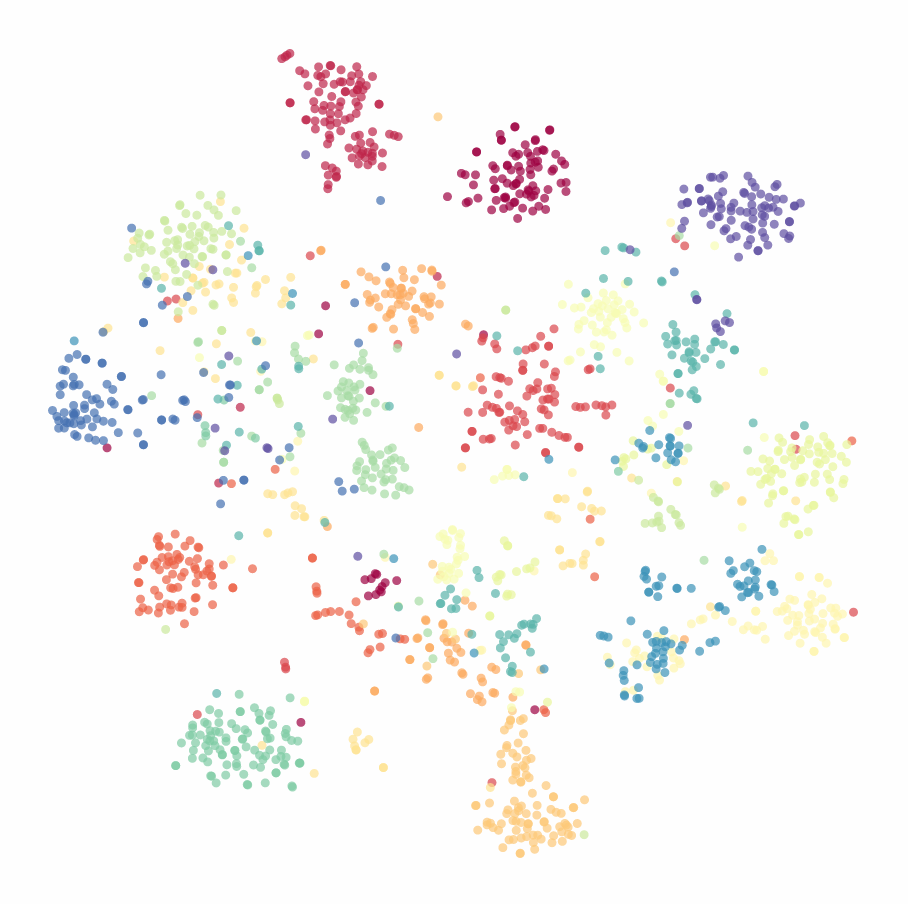}
    \caption{SH embeddings}
    \label{fig:tsne_clip_delta}
\end{subfigure}
\hfill
\begin{subfigure}[b]{0.245\textwidth}
    \includegraphics[width=\textwidth]{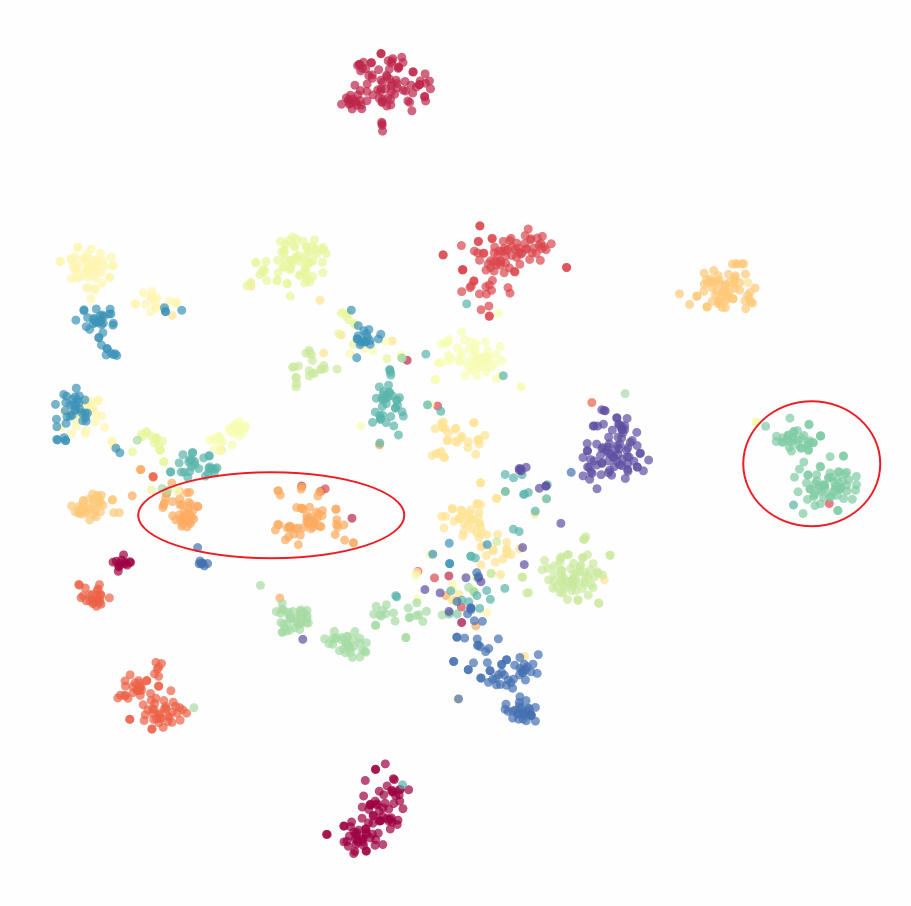}
    \caption{embeddings w/ \algo}
    \label{fig:tsne_ours}
\end{subfigure}
\hfill
\begin{subfigure}[b]{0.245\textwidth}
    \includegraphics[width=\textwidth]{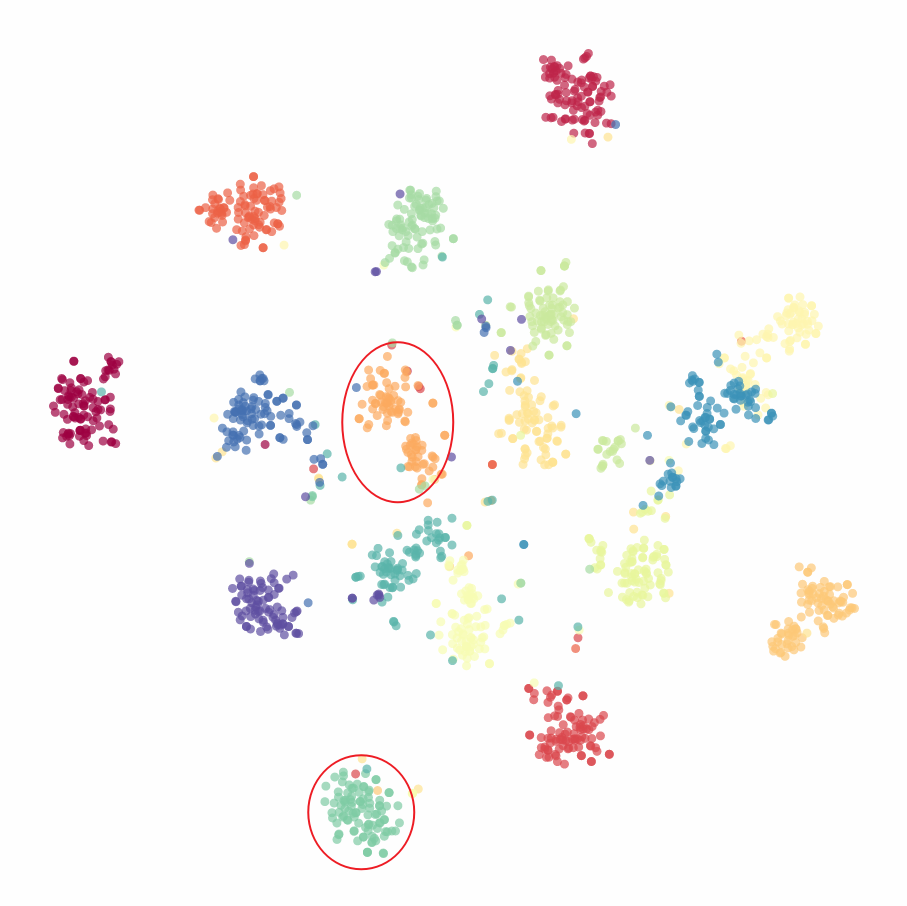}
    \caption{SH embeddings after \algo}
    \label{fig:tsne_ours_delta}
\end{subfigure}
\caption{The t-SNE visualization of embeddings on DomainNet. SH is an abbreviation for style-homogenized.}
\label{fig:tsne}
\end{figure*}

\begin{figure}[ht]
\centering
\includegraphics[width=1.0\columnwidth]{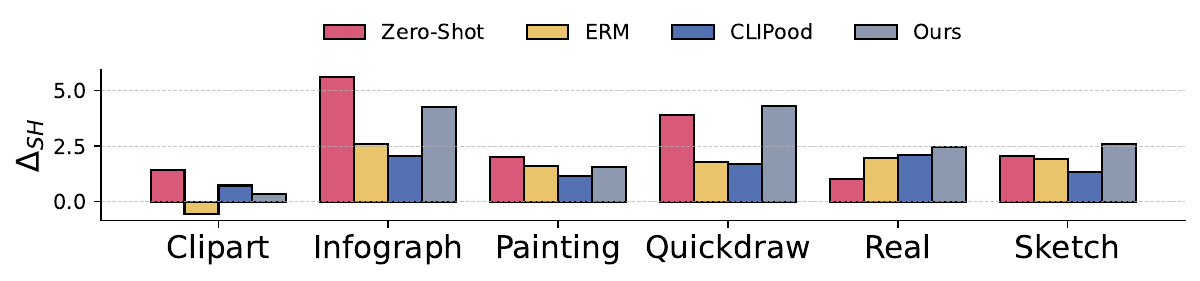}
\caption{Performance change $\Delta_{SH}$ before and after applying style-homogenized inference on DomainNet.}
\label{fig:delta_df}
\end{figure}

\subsection{Datasets and Evaluation Metrics}

To assess the effectiveness of our approach, we perform evaluations on five standard domain generalization benchmarks, including PACS \cite{li2017deeper}, VLCS \cite{torralba2011unbiased}, OfficeHome \cite{venkateswara2017deep}, TerraIncognita \cite{beery2018recognition}, and DomainNet \cite{peng2019moment}. PACS consists of 9,991 images from 7 object categories across 4 distinct domains (Photo, Art painting, Cartoon, and Sketch). VLCS comprises 10,729 images spanning 5 categories from 4 domains, drawn from multiple classical computer vision datasets. OfficeHome contains 15,579 images from 65 object categories in 4 domains (Art, Clipart, Product, and Real-World). TerraIncognita includes 24,788 wildlife images from 10 categories across 4 geographic domains, highlighting domain shifts due to environmental differences. DomainNet, the largest and most challenging benchmark, contains approximately 600,000 images from 345 categories in 6 diverse domains (Clipart, Infograph, Painting, Quickdraw, Real, and Sketch), introducing significant challenges in both domain and category shifts. The performance is measured by Top-1 classification accuracy (\%) on novel domain dataset $\mathcal{D}^{\mathrm{test}}$. Results are reported as the mean and standard deviation across three random seeds.

\subsection{Implementation Details}

Following previous method \cite{shu2023clipood}, we use pre-trained CLIP ViT-B/16 \cite{dosovitskiy2020image,radford2021learning} as backbone. The image encoder is full fine-tuned and the text encoder is frozen during training. In our experiments, the parameter $\tau$ in \Cref{eq:delta_align} was set to $\frac{1}{20}$ for the DomainNet dataset, and to $\frac{1}{5}$ for the other datasets. Unless otherwise specified, both temperature parameters $\tau_{\scriptscriptstyle\mathrm{SWM}}$ and $\tau_{\scriptscriptstyle{C}}$ were fixed at $\frac{1}{100}$ across all datasets. We employ the AdamW optimizer \cite{loshchilov2017decoupled} along with cosine learning rate schedule for all datasets. Additional experimental details can be found in supplementary material.


\subsection{Main Results}

We conduct extensive comparisons with state-of-the-art methods on five widely-used benchmark datasets to evaluate the effectiveness of \algo, and the results are shown in \Cref{tab:main_res}. We provide the results of CLIP zero-shot, style-homogenized zero-shot, and standard fine-tuning for CLIP pre-trained model (ERM). Across the majority of datasets, style-homogenized zero-shot demonstrates superior performance compared to CLIP zero-shot, yielding an average gain of around 0.84\%. We also compare our method with recent domain generalization approaches based on the pretrained CLIP model, including CLIPood \citep{shu2023clipood}, VLV2-SD \citep{addepalli2024leveraging}, and CLIPCEIL \citep{yu2024clipceil}. Overall, we argue that our results are significant, especially given the high difficulty of DG tasks. \algo\ consistently outperforms all competing methods across all five benchmarks. This consistency is a critical differentiator; in contrast, competing methods like CLIPooD and CLIPCEIL failed to show consistent or significant gains across all tested scenarios. Remarkably, \algo\ achieves an average performance gain of 2.7\% over ERM, with considerable gains of 4.4\% on OfficeHome and 4.0\% on DomainNet. It is particularly worth emphasizing that \algo\ surpasses the state-of-the-art CLIPCEIL by $0.5$ and $0.3$ on TerraIncognita and DomainNet, respectively. These datasets are known for their difficulty due to wild imagery and large category diversity, indicating the robustness of \algo\ in complex settings. Detailed results for each domain within datasets, along with further analysis, are provided in supplementary material.

\subsection{Analysis of \algo}
\noindent
\textbf{Ablation Analysis.}
To comprehensively understand the contribution of each component within our proposed \algo, we present a detailed ablation study in \Cref{tab:ablations}. Notably, replacing our style-homogenized alignment with image-text alignment loss during training leads to a significant degradation in performance across all domains. The most pronounced drop is observed in the ``Painting'' domain, which experiences a 1.5\% performance decrease. This evidence indicates that style-homogenized alignment is instrumental in bolstering the model's robustness and its capacity for generalization when encountering out-of-distribution domains. The removal of alignment regularization and prediction combination from the original CLIP model resulted in a performance drop of 0.4\% to 0.5\%, indicating that semantic support from CLIP facilitates the model's generalization across various complex domains. When removing additional centroids, the model suffered a performance degradation, with the ``Quickdraw'' domain experiencing the most pronounced drop of 0.7\%. Given the inherent difficulty of ``Quickdraw'' as a target domain \cite{kempf2025and}, this underscores the substantial contribution of additional centroids to the accurate prediction of stylistically anomalous samples. Furthermore, the removal of either CPM or SWM resulted in a measurable decrease in the model's generalization performance, which emphasizes the necessity of generating diverse additional centroids to generalize to more style domains.


\noindent
\textbf{Analysis on Performance Change.}
In \Cref{fig:delta_acc}, we present the performance changes of different methods relative to CLIP zero-shot on DomainNet. We observe that zero-shot performance using style-homogenized embeddings significantly outperforms original zero-shot, further demonstrating the effectiveness of style-homogenized embeddings in addressing domain shift. Moreover, \algo\ achieves the best performance, and the performance advantage is more pronounced on domains such as ``Infograph'' and ``Quickdraw'', which exhibit greater divergence from natural image styles and existing domain generalization methods always fail to achieve competitive performance under these circumstances.

\noindent
\textbf{Analysis on Additional Centroids.}
During the construction of additional centroids in \algo, a large number of domain-specific style words are utilized, covering common domain styles, which may potentially include the real style of the target domain. This highlights the importance of examining how real target domain word influences the model's performance. \Cref{fig:domain_word} shows the performance difference of \algo\ compared to its variant that constructs centroids without using real target domain style word. While the use of true domain-style terms leads to consistent improvements across most target domains, the overall performance gain is marginal, remaining under 0.1\%. This suggests that \algo\ does not rely on true domain-specific style words and can instead enhance model robustness through the use of a large and diverse set of domain words. Additionally, we observe that domains ``Infograph'' and ``Quickdraw'' are slightly more sensitive to the inclusion of real domain-specific words, which is primarily attributed to the significant disparity between these two domains and common domains. This offers flexibility to our method: if the approximate domain style is known especially for the rare styles, incorporating corresponding domain words into centroids construction can further improve generalization to the target domain.

\noindent
\textbf{Analysis on Embedding Visualizations.}
We visualize image and style-homogenized embeddings using t-SNE \cite{van2008visualizing} in \Cref{fig:tsne}. Comparing \Cref{fig:tsne_clip} and \Cref{fig:tsne_clip_delta} before training, we observe that style-homogenized embeddings exhibit better class separability than original embeddings, suggesting that style information in original embeddings may negatively impact classification performance. After training with \algo\, a comparison between \Cref{fig:tsne_ours} and \Cref{fig:tsne_ours_delta} (highlighted in red) reveals that style-homogenized embeddings form more compact class clusters than image embeddings, showing its robustness under domains with unusual or unseen styles.

\noindent
\textbf{Analysis on Domain-Agnostic Centroid Aggregation.}
To analyze the impact of domain-agnostic centroid aggregation inference on different methods, we present in \Cref{fig:delta_df} the performance differences compared to direct prediction using image-text embeddings. The results indicate that domain-agnostic centroid aggregation inference leads to more pronounced improvements for zero-shot and \algo\ on the majority of target domains, highlighting the effectiveness of \algo\ in preserving CLIP’s inherent representational structure, which is crucial for synthesizing style-homogenized embeddings. ERM and CLIPood exhibit relatively limited performance gains, which we attribute to the asymmetric image-text alignment that partially disrupts the representational structure of the original CLIP model, thereby leading to significant deviations in the construction of style-homogenized embeddings. 


\section{Conclusion}
This paper introduces \algo, a novel CLIP-based approach for domain generalization. \algo\ addresses the critical limitation of asymmetric information between images (containing class+style) and text (containing only class) by aligning style-homogenized embeddings instead of raw representations. It removes domain-specific style centroids from both modalities during training. For inference without target domain labels, \algo\ projects diverse text domain centroids into the visual space and aggregates predictions via soft assignment. Extensive experiments on five benchmarks demonstrate that \algo\ achieves state-of-the-art performance, significantly outperforming prior methods. We argue that \algo\ introduces a novel perspective that sheds light on how pre-trained models can be more effectively exploited for domain generalization tasks.






\bibliographystyle{named}
\bibliography{ijcai26}

@article{zhou2022domain,
  title={Domain generalization: A survey},
  author={Zhou, Kaiyang and Liu, Ziwei and Qiao, Yu and Xiang, Tao and Loy, Chen Change},
  journal={IEEE Transactions on Pattern Analysis and Machine Intelligence},
  volume={45},
  number={4},
  pages={4396--4415},
  year={2022},
  publisher={IEEE}
}

@inproceedings{shu2023clipood,
  title={Clipood: Generalizing clip to out-of-distributions},
  author={Shu, Yang and Guo, Xingzhuo and Wu, Jialong and Wang, Ximei and Wang, Jianmin and Long, Mingsheng},
  booktitle={International Conference on Machine Learning},
  pages={31716--31731},
  year={2023},
  organization={PMLR}
}

@inproceedings{li2018domain,
  title={Domain generalization with adversarial feature learning},
  author={Li, Haoliang and Pan, Sinno Jialin and Wang, Shiqi and Kot, Alex C},
  booktitle={Proceedings of the IEEE Conference on Computer Vision and Pattern Recognition},
  pages={5400--5409},
  year={2018}
}

@inproceedings{li2018deep,
  title={Deep domain generalization via conditional invariant adversarial networks},
  author={Li, Ya and Tian, Xinmei and Gong, Mingming and Liu, Yajing and Liu, Tongliang and Zhang, Kun and Tao, Dacheng},
  booktitle={Proceedings of the European Conference on Computer Vision (ECCV)},
  pages={624--639},
  year={2018}
}

@inproceedings{radford2021learning,
  title={Learning transferable visual models from natural language supervision},
  author={Radford, Alec and Kim, Jong Wook and Hallacy, Chris and Ramesh, Aditya and Goh, Gabriel and Agarwal, Sandhini and Sastry, Girish and Askell, Amanda and Mishkin, Pamela and Clark, Jack and others},
  booktitle={International Conference on Machine Learning},
  pages={8748--8763},
  year={2021},
  organization={PMLR}
}

@article{li2021align,
  title={Align before fuse: Vision and language representation learning with momentum distillation},
  author={Li, Junnan and Selvaraju, Ramprasaath and Gotmare, Akhilesh and Joty, Shafiq and Xiong, Caiming and Hoi, Steven Chu Hong},
  journal={Advances in Neural Information Processing Systems},
  volume={34},
  pages={9694--9705},
  year={2021}
}

@article{zhou2022learning,
  title={Learning to prompt for vision-language models},
  author={Zhou, Kaiyang and Yang, Jingkang and Loy, Chen Change and Liu, Ziwei},
  journal={International Journal of Computer Vision},
  volume={130},
  number={9},
  pages={2337--2348},
  year={2022},
  publisher={Springer}
}

@inproceedings{zhou2022conditional,
  title={Conditional prompt learning for vision-language models},
  author={Zhou, Kaiyang and Yang, Jingkang and Loy, Chen Change and Liu, Ziwei},
  booktitle={Proceedings of the IEEE/CVF Conference on Computer Vision and Pattern Recognition},
  pages={16816--16825},
  year={2022}
}

@article{wu2023clipself,
  title={Clipself: Vision transformer distills itself for open-vocabulary dense prediction},
  author={Wu, Size and Zhang, Wenwei and Xu, Lumin and Jin, Sheng and Li, Xiangtai and Liu, Wentao and Loy, Chen Change},
  journal={arXiv preprint arXiv:2310.01403},
  year={2023}
}

@inproceedings{gan2024erasing,
  title={Erasing the bias: fine-tuning foundation models for semi-supervised learning},
  author={Gan, Kai and Wei, Tong},
  booktitle={Proceedings of the 41st International Conference on Machine Learning},
  pages={14453--14470},
  year={2024}
}

@inproceedings{lai2023padclip,
  title={Padclip: Pseudo-labeling with adaptive debiasing in clip for unsupervised domain adaptation},
  author={Lai, Zhengfeng and Vesdapunt, Noranart and Zhou, Ning and Wu, Jun and Huynh, Cong Phuoc and Li, Xuelu and Fu, Kah Kuen and Chuah, Chen-Nee},
  booktitle={Proceedings of the IEEE/CVF International Conference on Computer Vision},
  pages={16155--16165},
  year={2023}
}

@inproceedings{feng2024rethinking,
  title={Rethinking domain adaptation and generalization in the era Of clip},
  author={Feng, Ruoyu and Yu, Tao and Jin, Xin and Yu, Xiaoyuan and Xiao, Lei and Chen, Zhibo},
  booktitle={2024 IEEE International Conference on Image Processing (ICIP)},
  pages={2585--2591},
  year={2024},
  organization={IEEE}
}

@article{yu2024clipceil,
  title={CLIPCEIL: Domain Generalization through CLIP via channel refinement and image-text aLignment},
  author={Yu, Xi and Yoo, Shinjae and Lin, Yuewei},
  journal={Advances in Neural Information Processing Systems},
  volume={37},
  pages={4267--4294},
  year={2024}
}

@inproceedings{peng2019moment,
  title={Moment matching for multi-source domain adaptation},
  author={Peng, Xingchao and Bai, Qinxun and Xia, Xide and Huang, Zijun and Saenko, Kate and Wang, Bo},
  booktitle={Proceedings of the IEEE/CVF International Conference on Computer Vision},
  pages={1406--1415},
  year={2019}
}

@inproceedings{shi2024long,
  title={Long-tail learning with foundation model: heavy fine-tuning hurts},
  author={Shi, Jiang-Xin and Wei, Tong and Zhou, Zhi and Shao, Jie-Jing and Han, Xin-Yan and Li, Yu-Feng},
  booktitle={International Conference on Machine Learning},
  pages={45014--45039},
  year={2024},
  organization={PMLR}
}

@inproceedings{chen2023center,
  title={Center-aware adversarial augmentation for single domain generalization},
  author={Chen, Tianle and Baktashmotlagh, Mahsa and Wang, Zijian and Salzmann, Mathieu},
  booktitle={Proceedings of the IEEE/CVF Winter Conference on Applications of Computer Vision},
  pages={4157--4165},
  year={2023}
}

@inproceedings{liangumfc,
  title={UMFC: Unsupervised multi-domain feature calibration for vision-language models},
  author={Liang, Jiachen and Hou, RuiBing and Hu, Minyang and Chang, Hong and Shan, Shiguang and Chen, Xilin},
  booktitle={The Thirty-eighth Annual Conference on Neural Information Processing Systems},
  year={2024}
}

@article{liang2022mind,
  title={Mind the gap: Understanding the modality gap in multi-modal contrastive representation learning},
  author={Liang, Victor Weixin and Zhang, Yuhui and Kwon, Yongchan and Yeung, Serena and Zou, James Y},
  journal={Advances in Neural Information Processing Systems},
  volume={35},
  pages={17612--17625},
  year={2022}
}

@inproceedings{cha2022domain,
  title={Domain generalization by mutual-information regularization with pre-trained models},
  author={Cha, Junbum and Lee, Kyungjae and Park, Sungrae and Chun, Sanghyuk},
  booktitle={European Conference on Computer Vision},
  pages={440--457},
  year={2022},
  organization={Springer}
}

@inproceedings{addepalli2024leveraging,
  title={Leveraging vision-language models for improving domain generalization in image classification},
  author={Addepalli, Sravanti and Asokan, Ashish Ramayee and Sharma, Lakshay and Babu, R Venkatesh},
  booktitle={Proceedings of the IEEE/CVF Conference on Computer Vision and Pattern Recognition},
  pages={23922--23932},
  year={2024}
}

@inproceedings{jia2021scaling,
  title={Scaling up visual and vision-language representation learning with noisy text supervision},
  author={Jia, Chao and Yang, Yinfei and Xia, Ye and Chen, Yi-Ting and Parekh, Zarana and Pham, Hieu and Le, Quoc and Sung, Yun-Hsuan and Li, Zhen and Duerig, Tom},
  booktitle={International Conference on Machine Learning},
  pages={4904--4916},
  year={2021},
  organization={PMLR}
}

@article{pham2023combined,
  title={Combined scaling for zero-shot transfer learning},
  author={Pham, Hieu and Dai, Zihang and Ghiasi, Golnaz and Kawaguchi, Kenji and Liu, Hanxiao and Yu, Adams Wei and Yu, Jiahui and Chen, Yi-Ting and Luong, Minh-Thang and Wu, Yonghui and others},
  journal={Neurocomputing},
  volume={555},
  pages={126658},
  year={2023},
  publisher={Elsevier}
}

@inproceedings{li2022blip,
  title={Blip: Bootstrapping language-image pre-training for unified vision-language understanding and generation},
  author={Li, Junnan and Li, Dongxu and Xiong, Caiming and Hoi, Steven},
  booktitle={International Conference on Machine Learning},
  pages={12888--12900},
  year={2022},
  organization={PMLR}
}

@inproceedings{li2023blip,
  title={Blip-2: Bootstrapping language-image pre-training with frozen image encoders and large language models},
  author={Li, Junnan and Li, Dongxu and Savarese, Silvio and Hoi, Steven},
  booktitle={International Conference on Machine Learning},
  pages={19730--19742},
  year={2023},
  organization={PMLR}
}

@inproceedings{houlsby2019parameter,
  title={Parameter-efficient transfer learning for NLP},
  author={Houlsby, Neil and Giurgiu, Andrei and Jastrzebski, Stanislaw and Morrone, Bruna and De Laroussilhe, Quentin and Gesmundo, Andrea and Attariyan, Mona and Gelly, Sylvain},
  booktitle={International Conference on Machine Learning},
  pages={2790--2799},
  year={2019},
  organization={PMLR}
}

@article{chen2022adaptformer,
  title={Adaptformer: Adapting vision transformers for scalable visual recognition},
  author={Chen, Shoufa and Ge, Chongjian and Tong, Zhan and Wang, Jiangliu and Song, Yibing and Wang, Jue and Luo, Ping},
  journal={Advances in Neural Information Processing Systems},
  volume={35},
  pages={16664--16678},
  year={2022}
}

@inproceedings{jia2022visual,
  title={Visual prompt tuning},
  author={Jia, Menglin and Tang, Luming and Chen, Bor-Chun and Cardie, Claire and Belongie, Serge and Hariharan, Bharath and Lim, Ser-Nam},
  booktitle={European Conference on Computer Vision},
  pages={709--727},
  year={2022},
  organization={Springer}
}

@article{dong2022lpt,
  title={Lpt: Long-tailed prompt tuning for image classification},
  author={Dong, Bowen and Zhou, Pan and Yan, Shuicheng and Zuo, Wangmeng},
  journal={arXiv preprint arXiv:2210.01033},
  year={2022}
}

@inproceedings{zhou2023ods,
  title={Ods: Test-time adaptation in the presence of open-world data shift},
  author={Zhou, Zhi and Guo, Lan-Zhe and Jia, Lin-Han and Zhang, Dingchu and Li, Yu-Feng},
  booktitle={International Conference on Machine Learning},
  pages={42574--42588},
  year={2023},
  organization={PMLR}
}

@article{osowiechi2024watt,
  title={WATT: Weight average test-time adaptation of CLIP},
  author={Osowiechi, David and Noori, Mehrdad and Hakim, Gustavo Adolfo Vargas and Yazdanpanah, Moslem and Bahri, Ali and Cheraghalikhani, Milad and Dastani, Sahar and Beizaee, Farzad and Ayed, Ismail Ben and Desrosiers, Christian},
  journal={arXiv preprint arXiv:2406.13875},
  year={2024}
}

@inproceedings{nam2021reducing,
  title={Reducing domain gap by reducing style bias},
  author={Nam, Hyeonseob and Lee, HyunJae and Park, Jongchan and Yoon, Wonjun and Yoo, Donggeun},
  booktitle={Proceedings of the IEEE/CVF Conference on Computer Vision and Pattern Recognition},
  pages={8690--8699},
  year={2021}
}

@article{segu2023batch,
  title={Batch normalization embeddings for deep domain generalization},
  author={Segu, Mattia and Tonioni, Alessio and Tombari, Federico},
  journal={Pattern Recognition},
  volume={135},
  pages={109115},
  year={2023},
  publisher={Elsevier}
}

@article{bui2021exploiting,
  title={Exploiting domain-specific features to enhance domain generalization},
  author={Bui, Manh-Ha and Tran, Toan and Tran, Anh and Phung, Dinh},
  journal={Advances in Neural Information Processing Systems},
  volume={34},
  pages={21189--21201},
  year={2021}
}

@inproceedings{li2017deeper,
  title={Deeper, broader and artier domain generalization},
  author={Li, Da and Yang, Yongxin and Song, Yi-Zhe and Hospedales, Timothy M},
  booktitle={Proceedings of the IEEE International Conference on Computer Vision},
  pages={5542--5550},
  year={2017}
}

@inproceedings{torralba2011unbiased,
  title={Unbiased look at dataset bias},
  author={Torralba, Antonio and Efros, Alexei A},
  booktitle={CVPR 2011},
  pages={1521--1528},
  year={2011},
  organization={IEEE}
}

@inproceedings{venkateswara2017deep,
  title={Deep hashing network for unsupervised domain adaptation},
  author={Venkateswara, Hemanth and Eusebio, Jose and Chakraborty, Shayok and Panchanathan, Sethuraman},
  booktitle={Proceedings of the IEEE Conference on Computer Vision and Pattern Recognition},
  pages={5018--5027},
  year={2017}
}

@inproceedings{beery2018recognition,
  title={Recognition in terra incognita},
  author={Beery, Sara and Van Horn, Grant and Perona, Pietro},
  booktitle={Proceedings of the European Conference on Computer Vision (ECCV)},
  pages={456--473},
  year={2018}
}

@article{dosovitskiy2020image,
  title={An image is worth 16x16 words: Transformers for image recognition at scale},
  author={Dosovitskiy, Alexey and Beyer, Lucas and Kolesnikov, Alexander and Weissenborn, Dirk and Zhai, Xiaohua and Unterthiner, Thomas and Dehghani, Mostafa and Minderer, Matthias and Heigold, Georg and Gelly, Sylvain and others},
  journal={arXiv preprint arXiv:2010.11929},
  year={2020}
}

@article{loshchilov2017decoupled,
  title={Decoupled weight decay regularization},
  author={Loshchilov, Ilya and Hutter, Frank},
  journal={arXiv preprint arXiv:1711.05101},
  year={2017}
}

@article{kempf2025and,
  title={When and how does CLIP enable domain and compositional generalization?},
  author={Kempf, Elias and Schrodi, Simon and Argus, Max and Brox, Thomas},
  journal={arXiv preprint arXiv:2502.09507},
  year={2025}
}

@article{van2008visualizing,
  title={Visualizing data using t-SNE.},
  author={Van der Maaten, Laurens and Hinton, Geoffrey},
  journal={Journal of Machine Learning Research},
  volume={9},
  number={11},
  year={2008}
}

@inproceedings{wen2025domain,
  title={Domain generalization in clip via learning with diverse text prompts},
  author={Wen, Changsong and Peng, Zelin and Huang, Yu and Yang, Xiaokang and Shen, Wei},
  booktitle={Proceedings of the IEEE/CVF Conference on Computer Vision and Pattern Recognition},
  pages={9559--9569},
  year={2025}
}

@inproceedings{luo2025lada,
    title={{LADA}: Scalable Label-Specific {CLIP} Adapter for Continual Learning},
    author={Mao-Lin Luo and Zi-Hao Zhou and Tong Wei and Min-Ling Zhang},
    booktitle={Forty-second International Conference on Machine Learning},
    year={2025}
}

@inproceedings{gan2025semi,
  title={Semi-supervised clip adaptation by enforcing semantic and trapezoidal consistency},
  author={Gan, Kai and Ye, Bo and Zhang, Min-Ling and Wei, Tong},
  booktitle={The Thirteenth International Conference on Learning Representations},
  year={2025}
}

@article{wei2026x,
  title={X-mahalanobis: Transformer feature mixing for reliable OOD detection},
  author={Wei, Tong and Wang, Bolin and Shi, Jiang-Xin and Li, Yu-Feng and Zhang, Min-Ling},
  journal={Advances in Neural Information Processing Systems},
  volume={38},
  pages={141958--141986},
  year={2026}
}

@article{wei2024vision,
  title={Vision-language models are strong noisy label detectors},
  author={Wei, Tong and Li, Hao-Tian and Li, Chun-Shu and Shi, Jiang-Xin and Li, Yu-Feng and Zhang, Min-Ling},
  journal={Advances in Neural Information Processing Systems},
  volume={37},
  pages={58154--58173},
  year={2024}
}

\clearpage

\appendix
\section*{Supplementary Material}


We provide supplementary material to facilitate a better understanding of the proposed \algo.

\begin{figure}[ht]
\centering
\begin{subfigure}[b]{0.43\textwidth}
    \includegraphics[width=\textwidth]{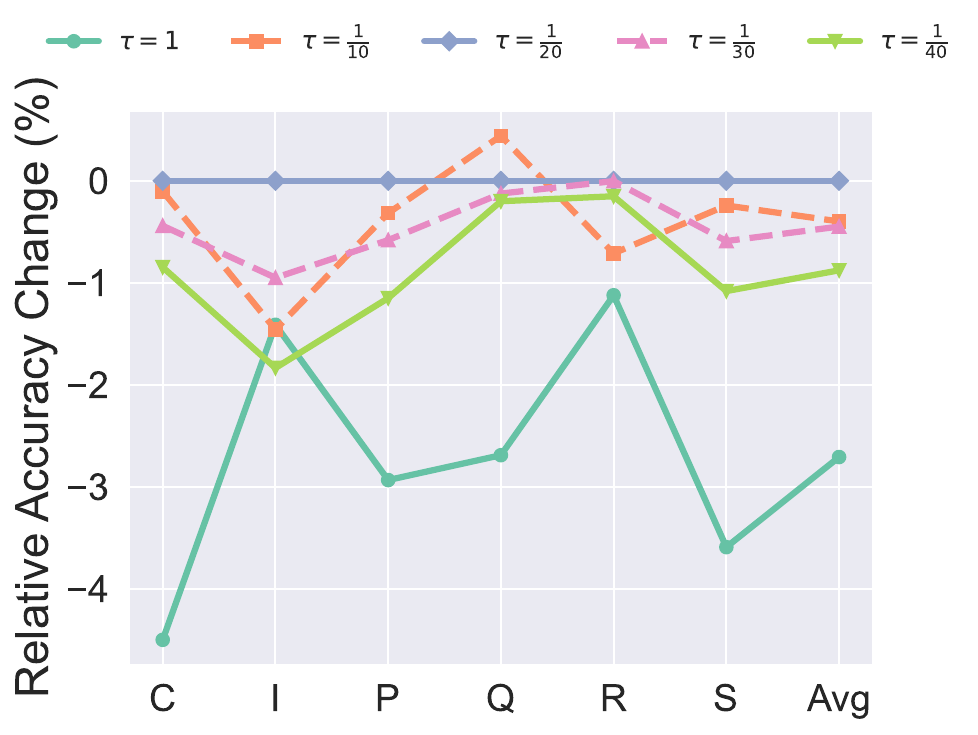}
    \caption{$\tau$ for domain-free alignment}
    \label{fig:sen_tau}
\end{subfigure}
\hfill
\begin{subfigure}[b]{0.46\textwidth}
    \includegraphics[width=\textwidth]{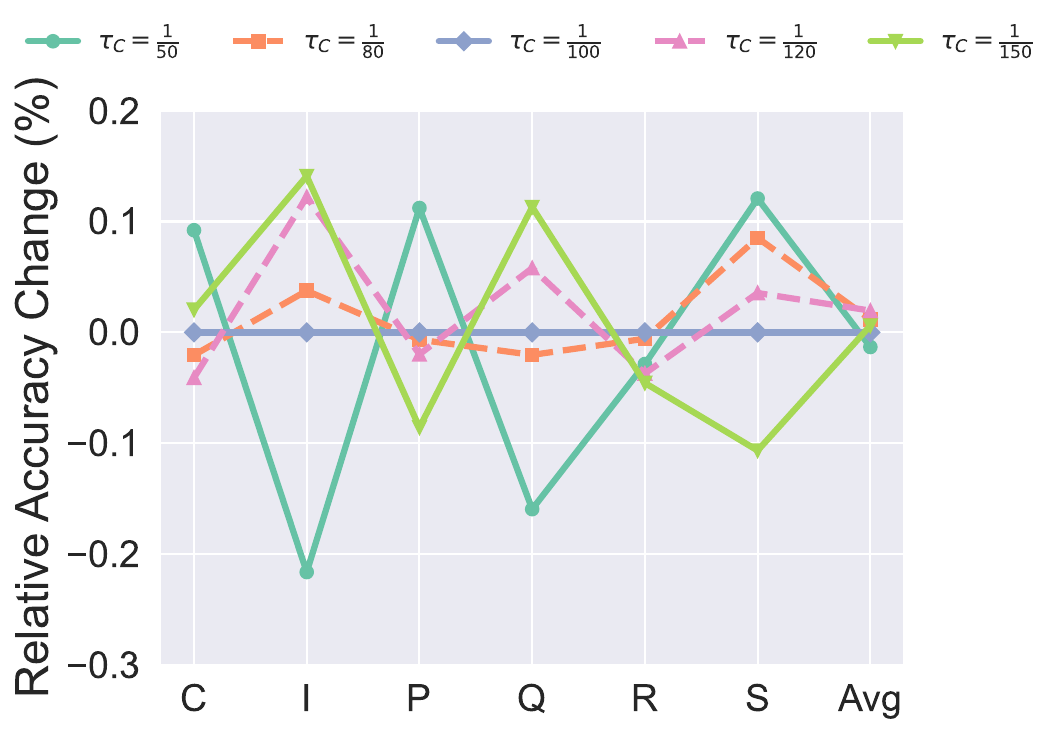}
    \caption{$\tau_C$ for soft assignment}
    \label{fig:sen_tauc}
\end{subfigure}
\caption{The sensitivity of $\tau$ and $\tau_C$ under various settings on DomainNet.}
\label{fig:sen}
\end{figure}

\section{More Implementation Details}
In this section, we provide additional implementation details of \algo. To ensure a fair comparison, we follow the most experimental setup of CLIPood. Given the relatively larger data volume of DomainNet, we train \algo\ for 20 epochs, whereas only 10 epochs are used for the other datasets. We train for 500 iterations per epoch. We employ the AdamW optimizer along with a cosine learning rate schedule for all datasets. We set the learning rate to $1 \times 10^{-5}$ for DomainNet and OfficeHome, and to $5 \times 10^{-6}$ for the other datasets. Due to the sensitivity of the model to alignment strength, we choose $\tau=\frac{1}{20}$ for DomainNet and $\tau=\frac{1}{5}$ for the remaining datasets. We hypothesize that the large scale of DomainNet necessitates a stronger alignment strength to maintain the model structure for learning domain-free features. The temperature parameters $\tau_{\scriptscriptstyle\mathrm{SWM}}$ is fixed at $\frac{1}{100}$ across all datasets. We set $\tau_{\scriptscriptstyle{C}}=\frac{1}{20}$ for VLCS and $\tau_{\scriptscriptstyle{C}}=\frac{1}{5}$ for TerraIncognita, while using $\tau_{\scriptscriptstyle{C}}=\frac{1}{100}$ for all other datasets. For each result of \algo, we report the average result and the standard deviation of three runs with random seeds.

\section{Templates for Additional Centroids}
We present in \Cref{tab:temp_add} the templates employed for constructing the additional centroids.

\begin{table}[htbp]
\centering
\begin{tabular}{@{}p{0.45\linewidth} p{0.45\linewidth}@{}}
\toprule
\multicolumn{2}{c}{\textbf{Template Prompt}} \\
\midrule
a drawing photo of a \{\} & a watercolor photo of a \{\} \\
a oil photo of a \{\} & a graffiti photo of a \{\} \\
a collage photo of a \{\} & a vector photo of a \{\} \\
a pixel photo of a \{\} & a minimalist photo of a \{\} \\
a abstract photo of a \{\} & a chart photo of a \{\} \\
a diagram photo of a \{\} & a blueprint photo of a \{\} \\
a surreal photo of a \{\} & a monochrome photo of a \{\} \\
a doodle photo of a \{\} & a clipart photo of a \{\} \\
a infograph photo of a \{\} & a painting photo of a \{\} \\
a quickdraw photo of a \{\} & a real photo of a \{\} \\
a sketch photo of a \{\} & a art photo of a \{\} \\
a product photo of a \{\} & a real world photo of a \{\} \\
a cartoon photo of a \{\} & a art painting photo of a \{\} \\
a white photo of a \{\} & a impressionist photo of a \{\} \\
a photorealistic photo of a \{\} & a vintage photo of a \{\} \\
a modern photo of a \{\} & a neon photo of a \{\} \\
a 3D photo of a \{\} & a pop art photo of a \{\} \\
a glitch photo of a \{\} & a isometric photo of a \{\} \\
a digital photo of a \{\} & an anime photo of a \{\} \\
a manga photo of a \{\} & a concept art photo of a \{\} \\
a futuristic photo of a \{\} & a cinematic photo of a \{\} \\
an abstract expressionism photo of a \{\} & an action painting photo of a \{\} \\
an art deco photo of a \{\} & an art nouveau photo of a \{\} \\
a baroque photo of a \{\} & a bauhaus photo of a \{\} \\
a cubism photo of a \{\} & an expressionism photo of a \{\} \\
a fauvism photo of a \{\} & an impressionism photo of a \{\} \\
a minimalism photo of a \{\} & a op art photo of a \{\} \\
a pointillism photo of a \{\} & a realism photo of a \{\} \\
a rococo photo of a \{\} & a surrealism photo of a \{\} \\
\bottomrule
\end{tabular}
\caption{Templates prompts for additional centroids.}
\label{tab:temp_add}
\end{table}

\begin{table}[ht]
\centering\small
\vspace{-0.08in}
\begin{adjustbox}{width=\linewidth}
\begin{tabular}{lccccc}
\toprule
Method & Art & Cartoon & Photo & Sketch & Avg \\
\midrule
CLIP Zero-Shot &
97.3 & 99.1 & 99.9 & 88.3 & 96.2 \\
CoOp &
98.3 & 98.8 & 99.7 & 87.3 & 96.0 \\
CoCoOp &
97.6 & 98.6 & 99.7 & 87.0 & 95.7 \\
MIRO &
- & - & - & - & 95.6 \\
CLIPood &
98.5 & 99.4 & 100.0 & 91.3 & 97.3 \\
CLIPCEIL &
- & - & - & - & 97.2 \\
\algo\ (ours) &
\textbf{98.9}\stdv{0.1} & \textbf{99.6}\stdv{0.0} & \textbf{100.0}\stdv{0.0} & \textbf{91.9}\stdv{0.1} & \textbf{97.6}\stdv{0.1} \\
\bottomrule
\end{tabular}
\end{adjustbox}
\caption{%
Comparison of classification accuracy on PACS. The best results are in \textbf{bold}.}
\label{tab:pacs_res}
\end{table}

\begin{table}[ht]
\centering\small
\vspace{-0.08in}
\begin{adjustbox}{width=\linewidth}
\begin{tabular}{lccccc}
\toprule
Method & Caltech & LabelMe & Sun & Pascal & Avg \\
\midrule
CLIP Zero-Shot &
98.9 & 65.5 & 77.6 & 84.5 & 81.7 \\
CoOp &
97.9 & 65.5 & 76.6 & 84.3 & 81.1 \\
CoCoOp &
99.8 & 67.0 & 78.5 & 87.1 & 83.1 \\
MIRO &
- & - & - & - & 82.2 \\
CLIPood &
100.0 & 67.7 & 80.2 & 92.1 & 84.8 \\
CLIPCEIL &
- & - & - & - & 85.2 \\
\algo\ (ours) &
\textbf{100.0}\stdv{0.0} & \textbf{68.3}\stdv{0.4} & \textbf{80.7}\stdv{0.2} & \textbf{92.4}\stdv{0.3} & \textbf{85.4}\stdv{0.2} \\
\bottomrule
\end{tabular}
\end{adjustbox}
\caption{%
Comparison of classification accuracy on VLCS. The best results are in \textbf{bold}.}
\label{tab:vlcs_res}
\end{table}

\section{Sensitivity Analysis}
\Cref{fig:sen} presents the results of a sensitivity analysis on several parameters within \algo. $\tau$ serves as the temperature parameter for domain-free alignment, and we set $\tau = \frac{1}{20}$ on DomainNet. As shown in \Cref{fig:sen_tau}, the performance of the model is negatively affected when the temperature parameter $\tau$ is set either too low or too high. We conjecture that a large $\tau$ value reduces the sharpness of the similarity distribution, thereby weakening the alignment and impeding the learning of class-discriminative features. When $\tau$ is set too low, the alignment becomes overly aggressive, which may lead to overfitting to noisy or uncertain image samples. Overall, setting $\tau = \frac{1}{20}$ achieves a good balance in the strength of domain-free alignment. Additional, as we vary $\tau_C$ during the soft assignment process, the results in \Cref{fig:sen_tauc} show no significant performance fluctuation, indicating that $\tau_c$ exhibits strong robustness.

\section{Full Results for Each Dataset}

\Cref{tab:pacs_res,tab:vlcs_res,tab:officehome_res,tab:terrainc_res,tab:domainnet_res} present the detailed results on the PACS, VLCS, OfficeHome, TerraIncognita, and DomainNet datasets. \algo\ achieves the best performance in most settings across each dataset. It is noteworthy that on datasets like PACS and VLCS, CoOp yields lower performance than CLIP zero-shot, indicating that such domain adaptation methods can significantly impair CLIP's inherent generalization capability. However, \algo\ exhibits consistently superior performance, with an average improvement of 9.2\% over CLIP zero-shot. We report the mean and standard deviation across three runs for \algo.

\begin{table}[ht]
\centering\small
\vspace{-0.08in}
\begin{adjustbox}{width=\linewidth}
\begin{tabular}{lccccc}
\toprule
Method & Art & Clipart & Product & Real & Avg \\
\midrule
CLIP Zero-Shot &
82.7 & 68.0 & 88.3 & 90.7 & 82.4 \\
CoOp &
82.8 & 69.7 & 91.0 & 90.6 & 83.5 \\
CoCoOp &
83.9 & 70.0 & 91.4 & 91.9 & 84.3 \\
MIRO &
- & - & - & - & 82.5 \\
CLIPood &
87.5 & 74.1 & 93.2 & \textbf{93.1} & 87.0 \\
CLIPCEIL &
- & - & - & - & 87.7 \\
\algo\ (ours) &
\textbf{88.3}\stdv{0.1} & \textbf{75.0}\stdv{0.1} & \textbf{93.9}\stdv{0.1} & 93.5\stdv{0.1} & \textbf{87.7}\stdv{0.1} \\
\bottomrule
\end{tabular}
\end{adjustbox}
\caption{%
Comparison of classification accuracy on OfficeHome. The best results are in \textbf{bold}.}
\label{tab:officehome_res}
\end{table}

\begin{table}[ht]
\centering\small
\vspace{-0.08in}
\begin{adjustbox}{width=\linewidth}
\begin{tabular}{lccccc}
\toprule
Method & L100 & L38 & L43 & L46 & Avg \\
\midrule
CLIP Zero-Shot &
51.2 & 23.4 & 29.9 & 29.1 & 33.4 \\
CoOp &
41.4 & 53.7 & 48.9 & 44.6 & 47.0 \\
CoCoOp &
50.7 & 56.0 & 51.9 & 44.0 & 50.4 \\
MIRO &
- & - & - & - & 54.3 \\
CLIPood &
73.9 & 63.6 & 57.5 & 46.6 & 60.4 \\
CLIPCEIL &
- & - & - & - & 62.0 \\
\algo\ (ours) &
\textbf{75.9}\stdv{0.5} & \textbf{66.4}\stdv{0.8} & \textbf{59.4}\stdv{0.3} & \textbf{48.4}\stdv{0.6} & \textbf{62.5}\stdv{0.3} \\
\bottomrule
\end{tabular}
\end{adjustbox}
\caption{%
Comparison of classification accuracy on TerraIncognita. The best results are in \textbf{bold}.}
\label{tab:terrainc_res}
\end{table}

\begin{table*}[ht]
\centering\small
\vspace{-0.08in}
\begin{adjustbox}{width=0.8\linewidth}
\begin{tabular}{lccccccc}
\toprule
Method & Clipart & Infograph & Painting & Quickdraw & Real & Sketch & Avg \\
\midrule
CLIP Zero-Shot &
71.3 & 47.4 & 66.4 & 14.2 & 83.4 & 63.1 & 57.5 \\
CoOp &
75.1 & 49.5 & 69.6 & 15.8 & 81.7 & 66.8 & 59.8 \\
CoCoOp &
74.8 & 51.9 & 69.2 & 16.0 & 80.9 & 67.2 & 60.0 \\
MIRO &
- & - & - & - & - & - & 54.0 \\
CLIPood &
77.8 & 54.5 & 72.6 & 20.2 & 85.6 & 70.1 & 63.5 \\
CLIPCEIL &
- & - & - & - & - & - & 63.6 \\
\algo\ (ours) &
\textbf{78.3}\stdv{0.1} & \textbf{55.0}\stdv{0.2} & \textbf{72.9}\stdv{0.2} & \textbf{21.0}\stdv{0.4} & \textbf{85.8}\stdv{0.1} & \textbf{70.3}\stdv{0.2} & \textbf{63.9}\stdv{0.1} \\
\bottomrule
\end{tabular}
\end{adjustbox}
\caption{%
Comparison of classification accuracy on DomainNet. The best results are in \textbf{bold}.}
\label{tab:domainnet_res}
\end{table*}

\section{Additional Experimental Analysis}

\subsection{Inference Efficiency}

To assess the practical applicability of \algo, we analyzed its computational cost during inference compared to the standard CLIP zero-shot method. The evaluation was performed on a single NVIDIA A100 GPU, measuring the average time to process one sample from the DomainNet dataset.

\begin{table}[h]
\centering\small
\begin{tabular}{lcc}
\toprule
\textbf{Method} & \textbf{Inference Time (ms/sample)} & \textbf{Overhead} \\
\midrule
Zero-Shot CLIP & 10.0 & - \\
\algo\ (Ours) & 10.8 & +8.0\% \\
\bottomrule
\end{tabular}
\caption{Comparison of inference time and computational overhead. The overhead is calculated relative to the CLIP zero-shot baseline.}
\label{tab:efficiency}
\end{table}

As shown in \Cref{tab:efficiency}, \algo\ introduces a minimal computational overhead of only 8.0\% compared to the standard CLIP zero-shot inference. This is because the additional steps in our pipeline, subtracting a centroid and aggregating predictions, primarily consist of highly optimized vector operations on pre-computed features. This marginal increase in computation demonstrates that \algo\ is a highly efficient framework, making it practical for deployment in real-world scenarios without significant performance trade-offs.

\subsection{Impact of Textual Centering}

We conducted an experiment to quantify the benefit of using diverse domain-styled prompts for textual centering. We compare our standard approach against a variant that computes the global text centroid using only a single generic prompt (i.e., ``a photo of a \{class name\}'').

\begin{table}[h]
\centering\small
\label{tab:textual_centering}
\begin{tabular}{lc}
\toprule
\textbf{Centering Strategy} & \textbf{DomainNet Accuracy (\%)} \\
\midrule
Single Generic Prompt & 63.6 \\
Diverse Domain-Styled Prompts & \textbf{63.9} \\
\bottomrule
\end{tabular}
\caption{Performance comparison on DomainNet using different textual centering strategies. Our approach utilizes diverse domain-styled prompts.}
\label{tab:textual_centering}
\end{table}

The results in \Cref{tab:textual_centering} show that using diverse domain-styled prompts yields a 0.3\% performance gain on the challenging DomainNet benchmark. This confirms that enriching the text modality with varied style cues helps create a more robust and representative global text centroid. This, in turn, mitigates potential biases introduced by a single, generic prompt and contributes to a more effective style-homogenized alignment.

\subsection{Ablation on Centroid Update Strategy}

In our method, the source domain centroids ($\mu_s$) are computed once before training and remain fixed. We compare this design choice against an alternative strategy where the centroids are updated dynamically during training using an Exponential Moving Average (EMA).

\begin{table}[h]
\centering
\begin{tabular}{lc}
\toprule
\textbf{Update Strategy} & \textbf{DomainNet Accuracy (\%)} \\
\midrule
EMA Update & 52.1 \\
Fixed Centroids (Ours) & \textbf{63.9} \\
\bottomrule
\end{tabular}
\caption{Performance on DomainNet with different centroid update strategies.}
\label{tab:ema_update}
\end{table}

\Cref{tab:ema_update} shows that using EMA to update centroids during training leads to a significant performance degradation of 11.8\% on DomainNet. We attribute this to the instability introduced by dynamically shifting centroids, which can disrupt the delicate process of aligning style-homogenized embeddings, especially in the early stages of training. In contrast, using fixed centroids computed from the powerful pre-trained CLIP encoder provides a stable anchor for style removal, preserving the model's geometric structure and leading to better generalization.

\subsection{Comparison with Tip-Adapter in Zero-Shot Setting}

To further situate our contribution, we compared our style-homogenized zero-shot inference mechanism with Tip-Adapter, a strong parameter-efficient fine-tuning method adapted for the zero-shot domain generalization task. For a fair comparison, both methods are applied directly on the pre-trained CLIP model without any training.

\begin{table}[h]
\centering\small
\begin{tabular}{lc}
\toprule
\textbf{Method} & \textbf{DomainNet Accuracy (\%)} \\
\midrule
Tip-Adapter (Zero-Shot DG) & 58.8 \\
Style-Homogenized Zero-Shot & \textbf{60.0} \\
\bottomrule
\end{tabular}
\caption{Comparison with Tip-Adapter on DomainNet in a zero-shot setting. Our method applies the domain-agnostic centroid aggregation inference on the original CLIP model.}
\label{tab:tip_adapter}
\end{table}

As detailed in \Cref{tab:tip_adapter}, our style-homogenized zero-shot approach outperforms Tip-Adapter by 1.2\% on DomainNet. While Tip-Adapter adapts to downstream data using a cached set of features, it does not explicitly account for domain style shifts. Our method, by directly modeling and mitigating style variations through centroid manipulation, proves to be a more effective mechanism for improving zero-shot generalization. This result further validates that explicitly addressing the style information in embeddings is a crucial component of our framework's success.

\section{Limitations and Future Work}
While \algo\ demonstrates strong performance, we acknowledge several limitations and avenues for future research. First, our method relies on a pre-defined set of textual domain prompts to construct synthetic centroids during inference. While we show this is robust, performance could potentially be further improved by developing methods to automatically generate or adapt these style prompts based on the test sample itself. Second, the effectiveness of style centering depends on the assumption that a domain's style bias can be reasonably captured by the centroid of its feature distribution. This may be less effective for domains with highly multi-modal style distributions. Future work could explore more sophisticated density estimators to model domain styles. Finally, applying and evaluating the \algo\ framework on other vision-language backbones and extending it to other DG tasks, such as semantic segmentation, represents a promising direction for future investigation.

\clearpage

\end{document}